\documentclass[10pt,twocolumn,letterpaper]{article}

\usepackage{cvpr}
\usepackage{times}
\usepackage{graphicx}
\usepackage{amsmath}
\usepackage{amssymb}
\usepackage{comment}
\usepackage{bbding}
\usepackage{booktabs}
\usepackage{diagbox}
\usepackage{siunitx}
\usepackage{pifont}
\newcommand{\cmark}{\ding{51}}%
\newcommand{\xmark}{\ding{55}}%
\usepackage[format=plain,labelfont=bf,up,textfont=up,font=small]{caption}
\usepackage{authblk}
\renewcommand\Affilfont{\fontsize{7.7}{14.4}\selectfont}

\usepackage{array}
\newcolumntype{L}[1]{>{\raggedright\let\newline\\\arraybackslash\hspace{0pt}}m{#1}}
\newcolumntype{C}[1]{>{\centering\let\newline\\\arraybackslash\hspace{0pt}}m{#1}}
\newcolumntype{R}[1]{>{\raggedleft\let\newline\\\arraybackslash\hspace{0pt}}m{#1}}

\newcommand\blfootnote[1]{%
  \begingroup
  \renewcommand\thefootnote{}\footnote{#1}%
  \addtocounter{footnote}{-1}%
  \endgroup
}
\usepackage[hang,flushmargin]{footmisc}
\makeatletter
\renewcommand\AB@affilsepx{  \hspace{1 mm}  \protect\Affilfont}
\makeatother

\usepackage[pagebackref=true,breaklinks=true,colorlinks,bookmarks=false]{hyperref}

\cvprfinalcopy 

\def\cvprPaperID{155} 

\ifcvprfinal\fi
\begin{document}

\title{Depth-Based 3D Hand Pose Estimation: \\ From Current Achievements to Future Goals}

\author{Shanxin Yuan$^1$\hspace*{4mm}Guillermo Garcia-Hernando$^1$\hspace*{4mm}Bj{\"o}rn Stenger$^2$\\Gyeongsik Moon$^4$\hspace*{4mm}Ju Yong Chang$^5$\hspace*{4mm}Kyoung Mu Lee$^4$\hspace*{4mm}Pavlo Molchanov$^6$\\Jan Kautz$^6$\hspace*{4mm}Sina Honari$^7$\hspace*{4mm}Liuhao Ge$^8$\hspace*{4mm}Junsong Yuan$^9$\hspace*{4mm} Xinghao Chen$^{10}$\hspace*{4mm}Guijin Wang$^{10}$\\Fan Yang$^{11}$\hspace*{4mm}Kai Akiyama$^{11}$\hspace*{4mm}Yang Wu$^{11}$\hspace*{4mm}Qingfu Wan$^{12}$\hspace*{4mm}Meysam Madadi$^{13}$\hspace*{4mm}Sergio Escalera$^{13,14}$\\Shile Li$^{15}$\hspace*{4mm}Dongheui Lee$^{15,16}$\hspace*{4mm}Iason Oikonomidis$^3$\hspace*{4mm}Antonis Argyros$^3$\hspace*{4mm}Tae-Kyun Kim$^1$}

\affil[]{}

\maketitle
\thispagestyle{empty}

\begin{abstract}

In this paper, we strive to answer two questions: What is the current state of 3D hand pose estimation from depth images? And, what are the next challenges that need to be tackled?
Following the successful \emph{Hands In the Million Challenge (HIM2017)}, we investigate the top 10 state-of-the-art methods on three tasks: single frame 3D pose estimation, 3D hand tracking, and hand pose estimation during object interaction.
We analyze the performance of different CNN structures with regard to hand shape, joint visibility, view point and articulation distributions. Our findings include:
(1)~isolated 3D hand pose estimation achieves low mean errors (10 mm) in the view point range of [70, 120] degrees, but it is far from being solved for extreme view points; 
(2)~3D volumetric representations outperform 2D CNNs, better capturing the spatial structure of the depth data;
(3)~Discriminative methods still generalize poorly to unseen hand shapes; 
(4)~While joint occlusions pose a challenge for most methods, explicit modeling of structure constraints can significantly narrow the gap between errors on visible and occluded joints.
\end{abstract}

\vspace{-5mm}
\section{Introduction}
\vspace{-1mm}
\blfootnote{
$^1$Imperial College London,
$^2$Rakuten Institute of Technology,
$^3$University of Crete and FORTH,
$^4$Seoul National University,
$^5$Kwangwoon University,
$^6$NVIDIA,
$^7$University of Montreal,
$^8$Nanyang Technological University,
$^9$State University of New York at Buffalo,
$^{10}$Tsinghua University,
$^{11}$Nara Institute of Science and Technology,
$^{12}$Fudan University,
$^{13}$Computer Vision Center,
$^{14}$University of Barcelona,
$^{15}$Technical University of Munich,
$^{16}$German Aerospace Center.\\ Corresponding author's email:  \href{mailto:s.yuan14@imperial.ac.uk}{\nolinkurl{s.yuan14@imperial.ac.uk}}
}
The field of 3D hand pose estimation has advanced rapidly, both in terms of accuracy~\cite{Seungryul2018, 
Choi_2017_learning,
choi2017robust, 
Camgoz_2017_ICCV,
deng2017hand3d,
mueller2017iccv,   
oberweger2016efficiently,
Remelli_2017_ICCV,
Simon_2017_CVPR,
tagliasacchi2015robust,
tang2015opening, 
tang2013real,
Wan_2017_CVPR, 
wan2016hand, 
ye2016spatial, 
zhang20163d,
Zimmermann_2017_ICCV} 
and dataset quality~\cite{garcia2017first,madadi2017occlusion, sun2015cascade, tang2017latentpami, tompson2014real, yuan2017bighand2}. 
Most successful methods treat the estimation task as a learning problem, using random forests or convolutional neural networks (CNNs). However, a review from 2015~\cite{supancic2015depth} surprisingly concluded that 
{\em a simple nearest-neighbor baseline outperforms most existing systems}. It concluded that {\em most systems do not generalize beyond their training sets}~\cite{supancic2015depth}, highlighting the need for more and better data.
Manually labeled datasets such as~\cite{qian2014realtime, handtracker_iccv2013} contain just a few thousand examples, making them unsuitable for large-scale training. Semi-automatic annotation methods, which combine manual annotation with tracking, help scaling the dataset size \cite{sun2015cascade, tang2014latent, tompson2014real}, but in the case of \cite{tang2014latent} the annotation errors are close to the lowest estimation errors. Synthetic data generation solves the scaling issue, but has not yet closed the realism gap, leading to some kinematically implausible poses~\cite{sharp2015accurate}. 

A recent study confirmed that cross-benchmark testing is poor due to different capture set-ups and annotation methods~\cite{yuan2017bighand2}. It showed that training a standard CNN on a million-scale dataset achieves state-of-the-art results. However, the estimation accuracy is not uniform, highlighting the well-known challenges of the task: variations in view point and hand shape, self-occlusion, and occlusion caused by objects being handled.

\begin{figure*}[t]
\begin{center}
		\includegraphics[width=1.0\textwidth]{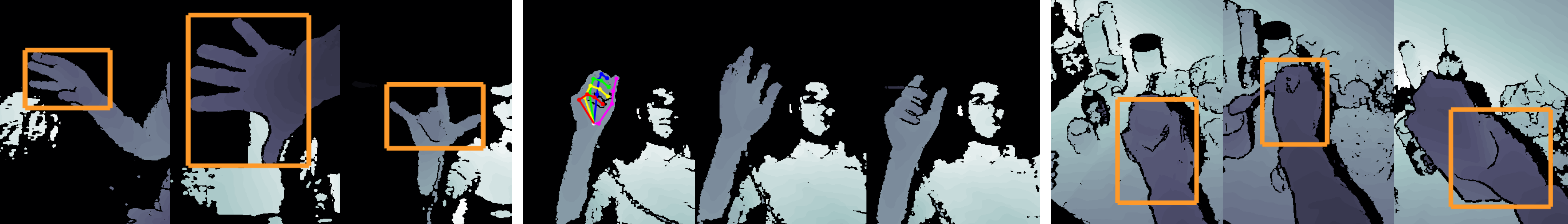}
    \caption{\textbf{Evaluated tasks}. For each scenario the goal is to infer the 3D locations of the 21 hand joints from a depth image. In \textbf{Single frame pose estimation} (left) and the \textbf{Interaction task} (right), each frame is annotated with a bounding box. In the \textbf{Tracking task} (middle), only the first frame of each sequence is fully annotated.}
	\label{pic:tasks}
	\vspace{-6mm}
\end{center}
\end{figure*}

In this paper we analyze the top methods of the {\em HIM2017} challenge \cite{HIM2017}. The benchmark dataset includes data from {\em BigHand2.2M} \cite{yuan2017bighand2} and the {\em First-Person Hand Action dataset (FHAD)} \cite{garcia2017first}, allowing the comparison of different algorithms in a variety of settings.
The challenge considers three different tasks: single-frame pose estimation, tracking, and hand-object interaction.
In the evaluation we consider different network architectures, preprocessing strategies, and data representations. Over the course of the challenge the lowest mean 3D estimation error could be reduced from  \SI{20}{\milli\meter}  to less than \SI{10}{\milli\meter}.
This paper analyzes the errors with regard to seen and unseen subjects, joint visibility, and view point distribution. 
We conclude by providing insights for designing the next generation of methods.

\paragraph{Related work.}
Public benchmarks and challenges in other areas such as ImageNet \cite{russakovsky2015imagenet} for scene classification and object detection, PASCAL \cite{everingham2015pascal} for semantic and object segmentation, and the VOT challenge \cite{kristan2015visual} for  object tracking, have been instrumental in driving progress in their respective field.
In the area of hand tracking, the review from 2007 by Erol \etal~\cite{erol2007vision} proposed a taxonomy of approaches. 
Learning-based approaches have been found effective for solving single-frame pose estimation, optionally in combination with hand model fitting for higher precision, \eg., \cite{taylor-siggraph2016}.
The review by Supancic \etal~\cite{supancic2015depth} compared 13 methods on a new dataset and concluded that deep models are well-suited to pose estimation~\cite{supancic2015depth}. It also highlighted the need for large-scale training sets in order to train models that generalize well. 
In this paper we extend the scope of previous analyses by comparing deep learning methods on a large-scale dataset, carrying out a fine-grained analysis of error sources and different design choices.

\section{Evaluation tasks}

We evaluate three different tasks on a dataset containing over a million annotated images using standardized evaluation protocols. 
Benchmark images are sampled from two datasets: {\em BigHand2.2M} \cite{yuan2017bighand2} and {\em First-Person Hand Action dataset (FHAD)} \cite{garcia2017first}. Images from \textit{BigHand2.2M} cover a large range of hand view points (including third-person and first-person views), articulated poses, and hand shapes. Sequences from the {\em FHAD} dataset are used to evaluate pose estimation during hand-object interaction.
Both datasets contain $640\times480$-pixel depth maps with 21 joint annotations, obtained from magnetic sensors and inverse kinematics.
The 2D bounding boxes have an average diagonal length of 162.4 pixels with a standard deviation of 40.7 pixels. 
The evaluation tasks are 3D single hand pose estimation, \ie, estimating the 3D locations of 21 joints, from
(1) individual frames,
(2) video sequences, given the pose in the first frame, and
(3) frames with object interaction, \eg, with a juice bottle, a salt shaker, or a milk carton. See Figure~\ref{pic:tasks} for an overview.
Bounding boxes are provided as input for tasks (1) and (3). 
The training data is sampled from the {\em BigHand2.2M} dataset and only the interaction task uses test data from the {\em FHAD} dataset.
See Table~\ref{tab:statistics} for dataset sizes and the number of total and unseen subjects for each task.

\begin{table}[t]
  \centering
  \small
  \begin{tabular}{lllll}
  \toprule 
   Number of   & Train & Test & Test   & Test  \\
  & & single & track & interact \\
  \midrule
  frames      & 957K  &  295K   & 294K   & 2,965 \\
  subjects (unseen)    & 5      &  10 (5)     & 10 (5)     & 2 (0)    \\
   \bottomrule  
   \end{tabular}
    \caption{\textbf{Data set sizes and number of subjects.}} 

  \label{tab:statistics}
\vspace{-3mm}
\end{table}

\section{Evaluated methods}

We evaluate the top 10 among 17 participating methods~\cite{HIM2017}.
Table~\ref{tab:methods} lists the methods with some of their key properties. We also indirectly evaluate \emph{DeepPrior} \cite{oberweger2015hands} and \emph{REN}~\cite{guo2017towards}, which are components of \emph{rvhand}~\cite{akiyama_rvhand17}, as well as \emph{DeepModel}~\cite{zhou2016model}, which is the backbone of \emph{LSL}~\cite{LSL17}.
We group methods based on different design choices.

\textbf{2D CNN \vs 3D CNN.} 
2D CNNs have been popular for 3D hand pose estimation~\cite{akiyama_rvhand17,chen2017pose,cai_Vanora17,guo2017towards,LSL17,madadi2017end,molchanov_Nvidia17,oberweger2015hands,ye2016spatial,zhou2016model}. 
Common pre-processing steps include cropping and resizing the hand volume by normalizing the depth values to [-1, 1]. 
Recently,  several methods have used a 3D CNN \cite{ge20173d,mks0601,Naisthand17}, where the input can be a 3D voxel grid \cite{mks0601,Naisthand17}, or a projective D-TSDF volume \cite{ge20173d}. Ge \emph{et al.}~\cite{gerobust} project the depth image onto three orthogonal planes and train a 2D CNN for each projection, then fusing the results. In \cite{ge20173d} they propose a 3D CNN by replacing 2D projections with a 3D volumetric representation (projective D-TSDF volumes~\cite{song2016deep}). In the \emph{HIM2017}  challenge~\cite{HIM2017}, they apply a 3D deep learning method~\cite{Osis17}, where the inputs are 3D points and surface normals. 
Moon \emph{et al.}~\cite{mks0601} propose a 3D CNN to estimate per-voxel likelihoods for each hand joint. 
\emph{NAIST\_RV}~\cite{Naisthand17} proposes a 3D CNN with a hierarchical branch structure, where the input is a 
$50^3$-voxel grid. 

\begin{table*}[t!]
\normalsize 
  \centering
  \resizebox{2.1\columnwidth}{!}{
  \begin{tabular}{lL{7cm}L{2cm}L{3cm}L{0.5cm}L{0.5cm}L{0.5cm}L{0.5cm}L{0.5cm}L{0.5cm}}
  \toprule 
  \bf Method   & \bf Model   & \bf Input  & \bf Aug. range (s,$\theta$,t)& \bf 3D & \bf De & \bf Hi &\bf St & \bf M &\bf R \\ 
  \midrule 

\emph{V2V-PoseNet}~\cite{mks0601}
& 3D CNN, per-voxel likelihood of each joint
& 88$\times$88$\times$88 voxels 
& [0.8, 1.2] [-40,40] [-8,8] 
&  \cmark 
&  \cmark 
&  \xmark 
&  \xmark
&  \xmark 
&  \cmark \\

\hline
\emph{RCN-3D}~\cite{molchanov_Nvidia17}
& RCN+ network \cite{honari2017improving} with 17 convolutional layers
& 80$\times$80 
& [0.7, 1.1] [0, 360] [-8,8]
& \xmark 
& \cmark 
& \xmark
& \xmark
& \cmark 
& \cmark \\

\hline
\emph{oasis}~\cite{Osis17}
& Hierarchical PointNet with three set abstraction levels and three full-connected layers
& 1024 3D points 
& random scaling* 
& \xmark 
& \xmark 
& \xmark 
& \cmark 
& \xmark 
& \xmark \\

\hline
\emph{THU\_VCLab}~\cite{chen2017pose}
& Pose-REN \cite{chen2017pose}: \emph{REN}~\cite{guo2017towards} + cascaded + hierarchical.
& 96$\times$96 
& [0.9,1.1] [-45,45]  [-5,5] 
& \xmark 
& \xmark
& \cmark
& \xmark 
& \cmark 
& \cmark \\

\hline
\emph{NAIST\_RV}~\cite{Naisthand17}
& 3D CNN with 5 branches, one for each finger 
& 50$\times$50$\times$50 3D grid
&  [0.9,1.1]  [-90, 90] [-15,15] 
& \cmark
& \xmark
& \cmark
& \xmark 
& \xmark
& \xmark  \\

\hline
\emph{Vanora}~\cite{cai_Vanora17}
& shallow CNN trained end-to-end
& resized 2D 
& random scaling*  
& \xmark
& \xmark 
& \xmark 
& \xmark 
& \xmark
& \xmark  \\

\hline
\emph{strawberryfg}~\cite{Strawberry} 
& ResNet-152 + \cite{sun2017compositional}   
& 224$\times$224  
& None  
& \xmark 
& \xmark 
& \xmark
& \cmark 
& \xmark 
& \cmark \\

\hline
\emph{rvhand}~\cite{akiyama_rvhand17} 
& ResNet \cite{he2016deep} + \emph{REN}~\cite{guo2017towards} + Deep Prior \cite{oberweger2015hands}  
& 192$\times$192  
&  [0.9,1.1]  [-90, 90] [-15,15] 
& \xmark 
& \xmark
& \cmark
& \cmark 
& \xmark 
& \cmark \\

\hline

\emph{mmadadi}~\cite{madadi2017end} 
& Hierarchical tree-like structured CNN~\cite{madadi2017end}
& 192$\times$192  
&  [random]  [-30, 30] [-10,10] 
& \xmark 
& \xmark 
& \cmark
& \cmark 
& \xmark 
& \xmark \\

\hline
 
\emph{LSL}~\cite{LSL17}
& ScaleNet to estimate hand scale + \emph{DeepModel}~\cite{zhou2016model} 
& 128$\times$128 
& [0.85,1.15]  [0,360]  [-20,20] 
& \xmark
& \xmark
& \xmark
& \cmark 
& \xmark
& \xmark \\

  \bottomrule
  \end{tabular}}
  
    \caption{\textbf{Methods evaluated in the hand pose estimation challenge.} Methods are ordered by average error on the leader-board. \newline
* in both methods, hand segmentation is performed considering different hand arm lengths. \textbf{3D}, \textbf{De}, \textbf{Hi}, \textbf{St}, \textbf{M}, and \textbf{R} denote \textbf{3D CNN}, \textbf{Detection-based method}, \textbf{Hierarchical model}, \textbf{Structure model}, \textbf{Multistage model}, and \textbf{Residual net}, respectively.}
  \label{tab:methods} 
  \vspace{-5mm} 
\end{table*}

\textbf{Detection-based \vs Regression-based.}
Detection-based methods~\cite{molchanov_Nvidia17,mks0601} 
produce a probability density map for each joint.
The method of \emph{RCN-3D}~\cite{molchanov_Nvidia17} is an RCN+ network \cite{honari2017improving}, based on Recombinator Networks (RCN) \cite{honari2016recombinator} with 17 layers and 64 output feature maps for all layers except the last one, which outputs a probability density map for each of the 21 joints. \emph{V2V-PoseNet}~\cite{mks0601} uses a 3D CNN to estimate per-voxel likelihood of each joint, and a CNN to estimate the center of mass from the cropped depth map. For training, 3D likelihood volumes are generated by placing normal distributions at the locations of hand joints.  
Regression-based methods~
\cite{akiyama_rvhand17,chen2017pose,Osis17, cai_Vanora17,LSL17,madadi2017end, oberweger2015hands,Naisthand17}
directly map the depth image to the joint locations or the joint angles  of a hand model \cite{sinha2016deephand,zhou2016model}. 
\emph{rvhand}~\cite{akiyama_rvhand17} combines ResNet~\cite{he2016deep}, Region Ensemble Network (\emph{REN})~\cite{guo2017towards}, and \emph{DeepPrior}~\cite{oberweger2015hands} to directly estimate the joint locations. 
\emph{LSL}~\cite{LSL17} uses one network to estimate a global scale factor and a second network~\cite{zhou2016model} to estimate all joint angles, which are fed into a forward kinematic layer to estimate the hand joints. 

\textbf{Hierarchical models} divide the pose estimation problem into sub-tasks~\cite{akiyama_rvhand17,chen2017pose,guo2017towards,madadi2017end,Naisthand17}.
The evaluated methods divide the hand joints either by finger ~\cite{madadi2017end,Naisthand17}, or by joint type~\cite{akiyama_rvhand17,chen2017pose,guo2017towards}. \emph{mmadadi}~\cite{madadi2017end} designs a hierarchically structured CNN, dividing the convolution+ReLU+pooling blocks into six branches (one per finger with palm and one for palm orientation), each of which is then followed by a fully connected layer. The final layers of all branches are concatenated into one layer to predict all joints. \emph{NAIST\_RV}~\cite{Naisthand17} chooses a similar hierarchical structure of a 3D CNN, but uses five branches, each to predict one finger and the palm.
\emph{THU\_VCLab}~\cite{chen2017pose}, \emph{rvhand}~\cite{akiyama_rvhand17}, and \emph{REN}~\cite{guo2017towards} apply constraints per finger and joint-type (across fingers) in their multiple regions extraction step, each region containing a subset of joints.  
All regions are concatenated in the last fully connected layers to estimate the hand pose.

\textbf{Structured methods} embed physical hand motion constraints into the model~\cite{Osis17,LSL17,madadi2017end,oberweger2015hands,Strawberry,zhou2016model}. Structural constraints are included in the CNN model~\cite{LSL17,oberweger2017deepprior,oberweger2015hands} or in the loss function \cite{madadi2017end,Strawberry}.
\emph{DeepPrior} \cite{oberweger2015hands} learns a prior model and integrates it into the network by introducing a \emph{bottleneck} in the last CNN layer. 
\emph{LSL}~\cite{LSL17} uses prior knowledge in \emph{DeepModel}~\cite{zhou2016model} by embedding a kinematic model layer into the CNN and using a fixed hand model. 
\emph{mmadadi} \cite{madadi2017end} includes the structure constraints in the loss function, which incorporates physical constraints about natural hand motion and deformation. 
\emph{strawberryfg}~\cite{Strawberry} applies a structure-aware regression approach, Compositional Pose Regression~\cite{sun2017compositional}, and replaces the original ResNet-50 with ResNet-152. It uses phalanges instead of joints for representing pose, and defines a loss function that encodes long-range interaction between the phalanges. 

\textbf{Multi-stage methods} propagate results from each stage to enhance the training of the subsequent stages~\cite{chen2017pose,molchanov_Nvidia17}.  \emph{THU\_VCLab}~\cite{chen2017pose} uses \emph{REN}~\cite{guo2017towards} to predict an initial hand pose. In the following stages, feature maps are computed with the guidance of the hand pose estimate in the previous stage. 
\emph{RCN-3D}~\cite{molchanov_Nvidia17} has five stages: (1) 2D landmark estimation using an RCN+ network~\cite{honari2017improving}, (2) estimation of corresponding depth values by multiplying probability density maps with the input depth image, 
(3) inverse perspective projection of the depth map to 3D, (4) error compensation for occlusions and depth errors (a 3-layer network of residual blocks) and, (5) error compensation for noise (another 3-layer network of residual blocks).

\textbf{Residual networks.}
ResNet~\cite{he2016deep} is adopted by several methods~\cite{akiyama_rvhand17, chen2017pose,guo2017towards, molchanov_Nvidia17, mks0601, sun2017compositional}. \emph{V2V-PoseNet}~\cite{mks0601} uses residual blocks as main building blocks. 
\emph{strawberryfg}~\cite{Strawberry} implements the Compositional Pose Regression method \cite{sun2017compositional} by using ResNet-152 as basic network. 
\emph{RCN-3D}~\cite{molchanov_Nvidia17} uses two small residual blocks in its fourth and fifth stage.

\vspace{-1mm} 
\section{Results}
\vspace{-1mm} 

The aim of this evaluation is to identify success cases and failure modes. We use both standard error metrics~\cite{oikonomidis2011efficient,sharp2015accurate,taylor2012vitruvian} and new proposed metrics to provide further insights. We consider joint visibility, seen \vs unseen subjects, hand view point distribution, articulation distribution, and per-joint accuracy.\\

\begin{figure}[t]
\begin{center}
        \includegraphics[trim=7.5cm 5cm 11cm 7cm, clip=true,width=0.8\columnwidth]{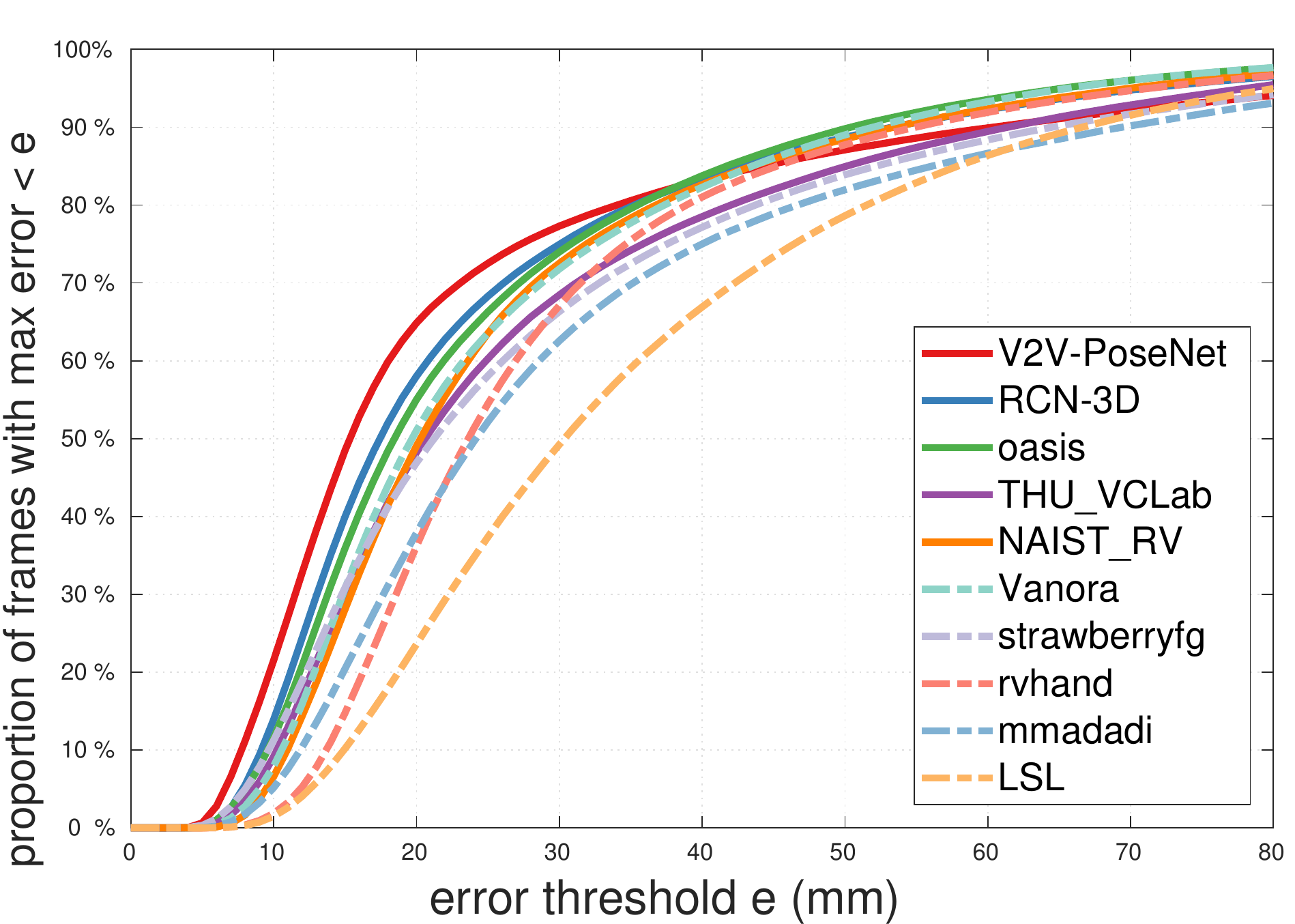}
        \includegraphics[trim=7.5cm 5cm 11cm 7cm, clip=true,width=0.8\columnwidth]{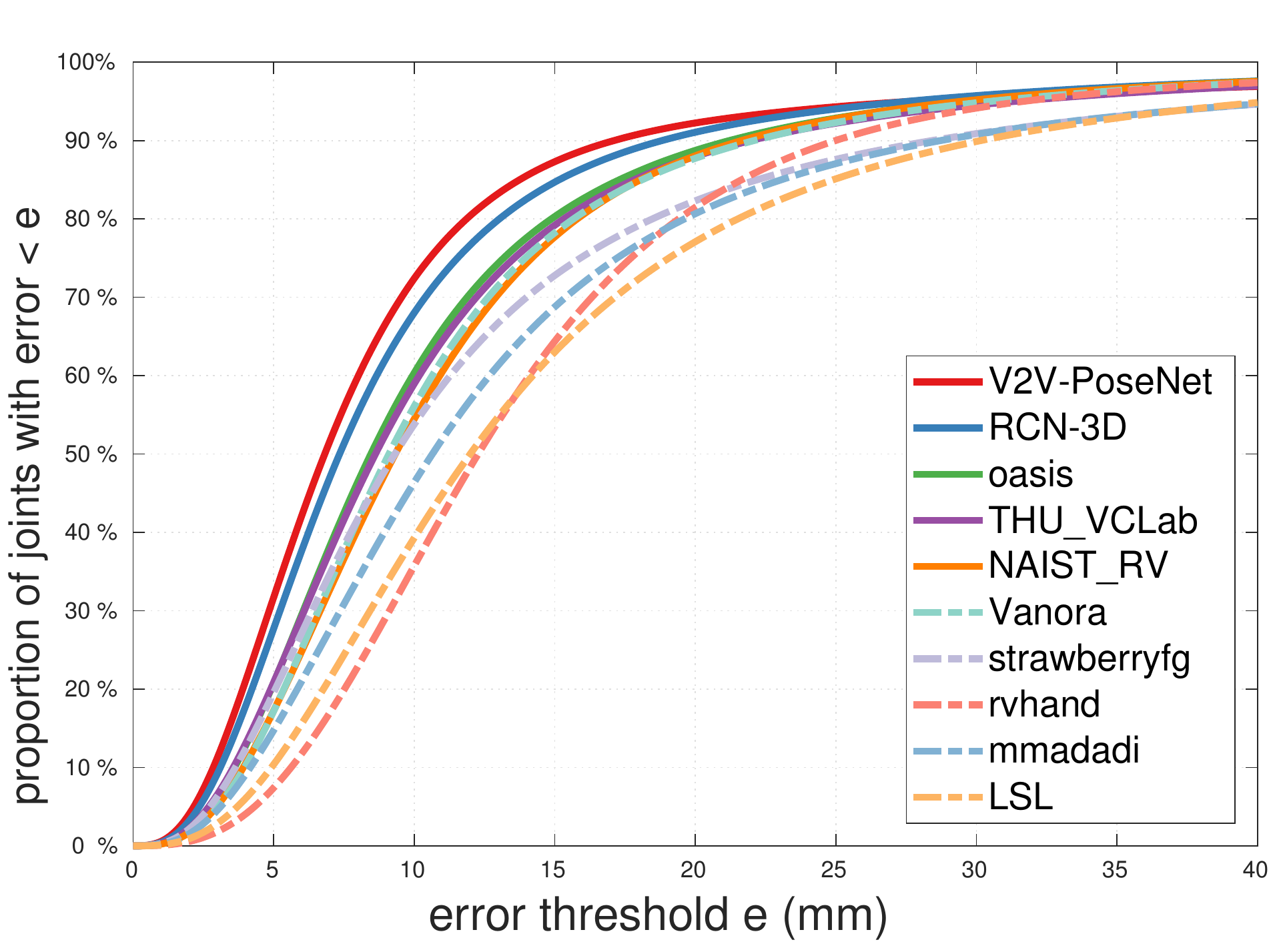}
        \hfill \\ 
         \caption{\textbf{Estimation errors.} (top) proportion of frames within maximum error threshold~\cite{taylor2012vitruvian}, (bottom) proportion of joints within an error threshold~\cite{sharp2015accurate}.}	\label{pic:frame_compare_methods_twometrics}
	\vspace{-5mm} 
\end{center}
\end{figure}

\begin{table}[t]
\small
  \centering
  
  \begin{tabular}{lR{1cm}R{1.1cm}R{1.1cm}R{1.2cm}}
  \toprule 
  \small
  \diagbox[width=2cm]{Method}{Case}     &  Seen Visible & Seen Occ & Unseen Visible  & Unseen Occ  \\
      \midrule 
V2V-PoseNet & {\bf 6.2} & {\bf 8.0} & 11.1  &  {\bf 14.6}  \\ 
RCN-3D & 6.9 & 9.0 & {\bf 10.6}    & 14.8  \\ 
oasis & 8.2  & 9.8  & 12.4  &  14.9  \\ 
THU\_VCLab & 8.4  & 10.2  & 12.5  &  16.1  \\ 
NAIST\_RV & 8.8  & 10.1  & 13.1  &  15.6  \\ 
Vanora & 8.8  & 10.5  & 12.9  &  15.5  \\ 
strawberryfg & 9.3  & 10.7  & 16.4  &  18.8  \\ 
rvhand & 12.2  & 11.9  & 16.1 &  17.6  \\ 
mmadadi & 10.6  & 13.6  & 15.6  &  19.7  \\ 
LSL & 11.8  & 13.1  & 18.1  &  19.2  \\ 
\midrule
Top5 & 7.7 & 9.4  & 11.9 &  15.2 \\ 
All  & 9.1 & 10.7 & 13.9 &  16.7 \\
  \bottomrule
  \end{tabular}
    \caption{\textbf{Mean errors (in mm) for single frame pose estimation, divided by cases.} `Seen' and `Unseen' refers to whether or not the hand shape was in the training set, and `Occ' denotes `Occluded joints'.}
  \label{tab:avgerrorframe} 
  \vspace{-3mm}  
\end{table}

\vspace{-3mm} 
\subsection{Single frame pose estimation}
\vspace{-1mm} 

Over the 6-week period of the challenge
the lowest mean error could be reduced from \SI{19.7}{\milli\meter} to \SI{10.0}{\milli\meter} by exploring new model types and improving data augmentation, optimization and initialization. 
For hand shapes seen during training, the mean error was reduced from \SI{14.6}{\milli\meter} to \SI{7.0}{\milli\meter}, and for unseen hand shapes from  \SI{24.0}{\milli\meter}  to \SI{12.2}{\milli\meter}.
Considering typical finger widths of 10-\SI{20}{\milli\meter}, these methods are becoming applicable to scenarios like pointing or motion capture, but may still lack sufficient accuracy for fine manipulation that is critical in some UI interactions.

We evaluate ten state-of-the-art methods (Table~\ref{tab:methods}) directly and three methods indirectly, which were used as components of others, \emph{DeepPrior}~\cite{oberweger2015hands}, \emph{REN}~\cite{guo2017towards}, and \emph{DeepModel}~\cite{zhou2016model}.  Figure~\ref{pic:frame_compare_methods_twometrics} shows the results in terms of two metrics: (top) the proportion of frames in which all joint errors are below a threshold~\cite{taylor2012vitruvian} and (bottom) the total proportion of joints below an error threshold~\cite{sharp2015accurate}.

\begin{figure}[t]
\begin{center}	
\includegraphics[trim=1.4cm 2.3cm 0cm 1.8cm, clip=true,width=0.48\textwidth]{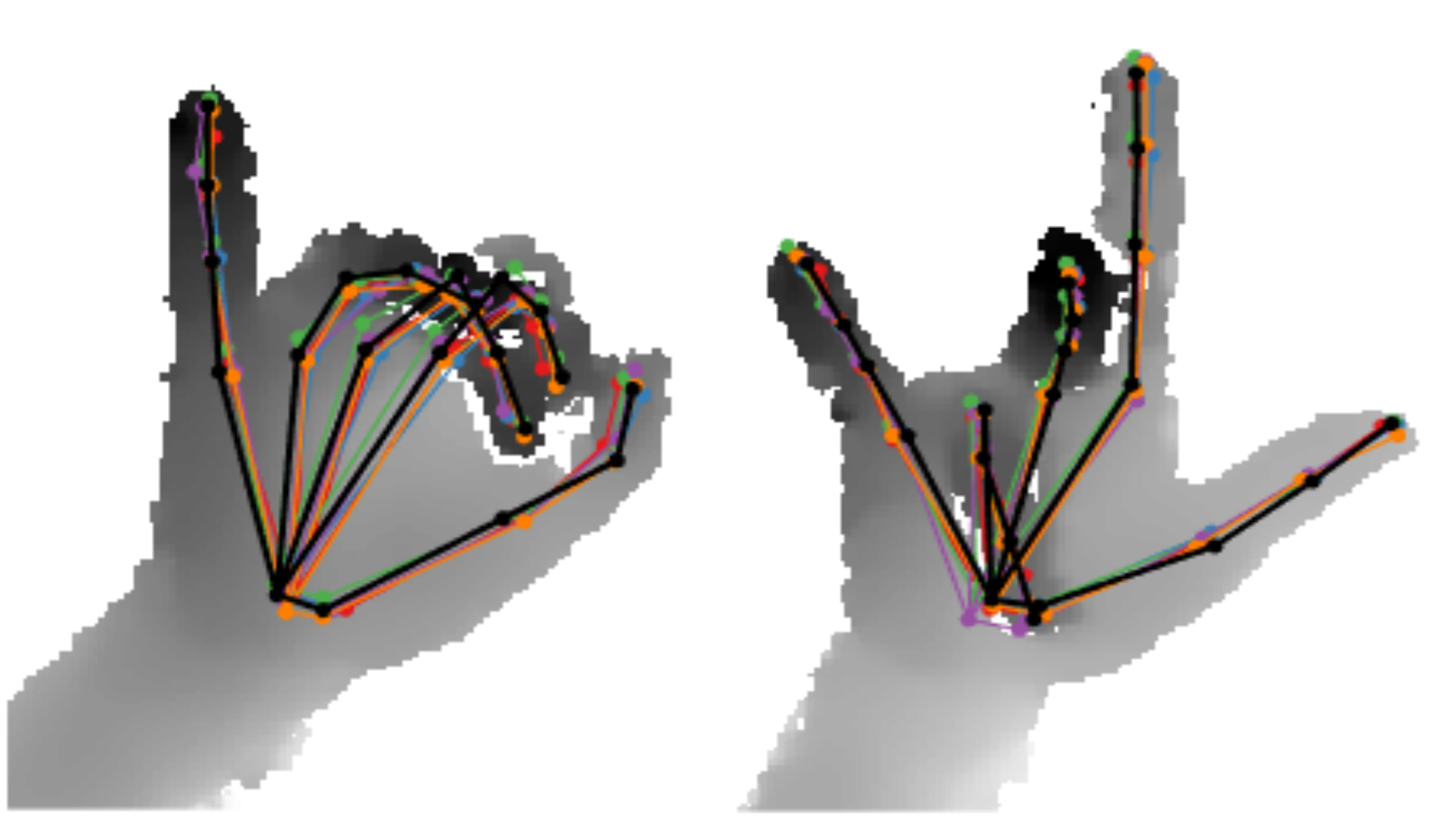}    \caption{\textbf{Estimating annotation errors.} Two examples of \emph{Easy Poses} overlayed with estimates by the top five methods (shown in different colors). Poses vary slightly, but are close to the ground truth (black).}    \label{pic:evalmetric_Uncertain} 
\vspace{-8mm} 
\end{center}
\end{figure}

\begin{figure*}[t]
\begin{center}
        \includegraphics[trim=7.5cm 5cm 11cm 7cm, clip=true,width=0.33\textwidth]{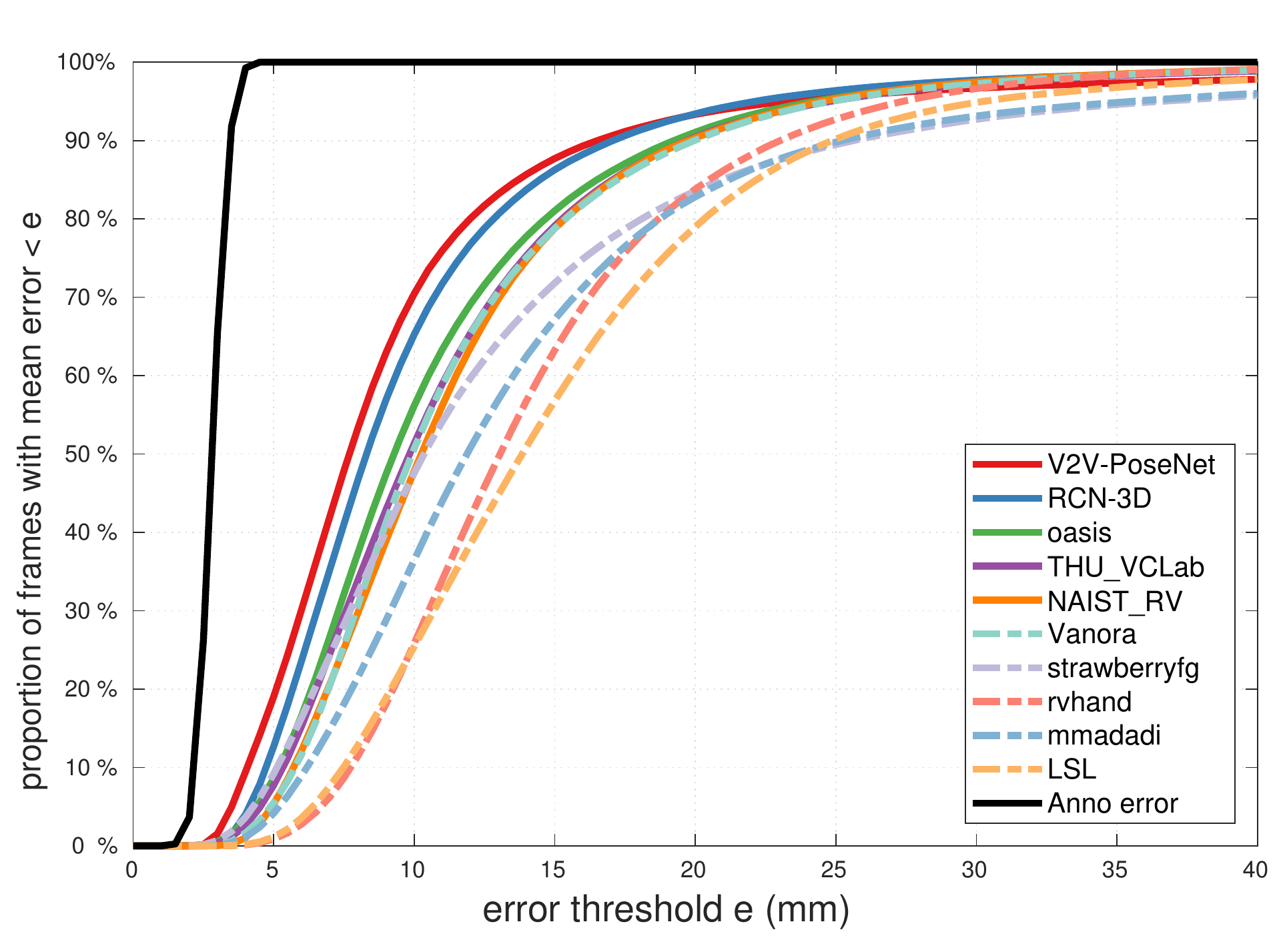}        
        \includegraphics[trim=7.5cm 5cm 11cm 7cm, clip=true,width=0.33\textwidth]{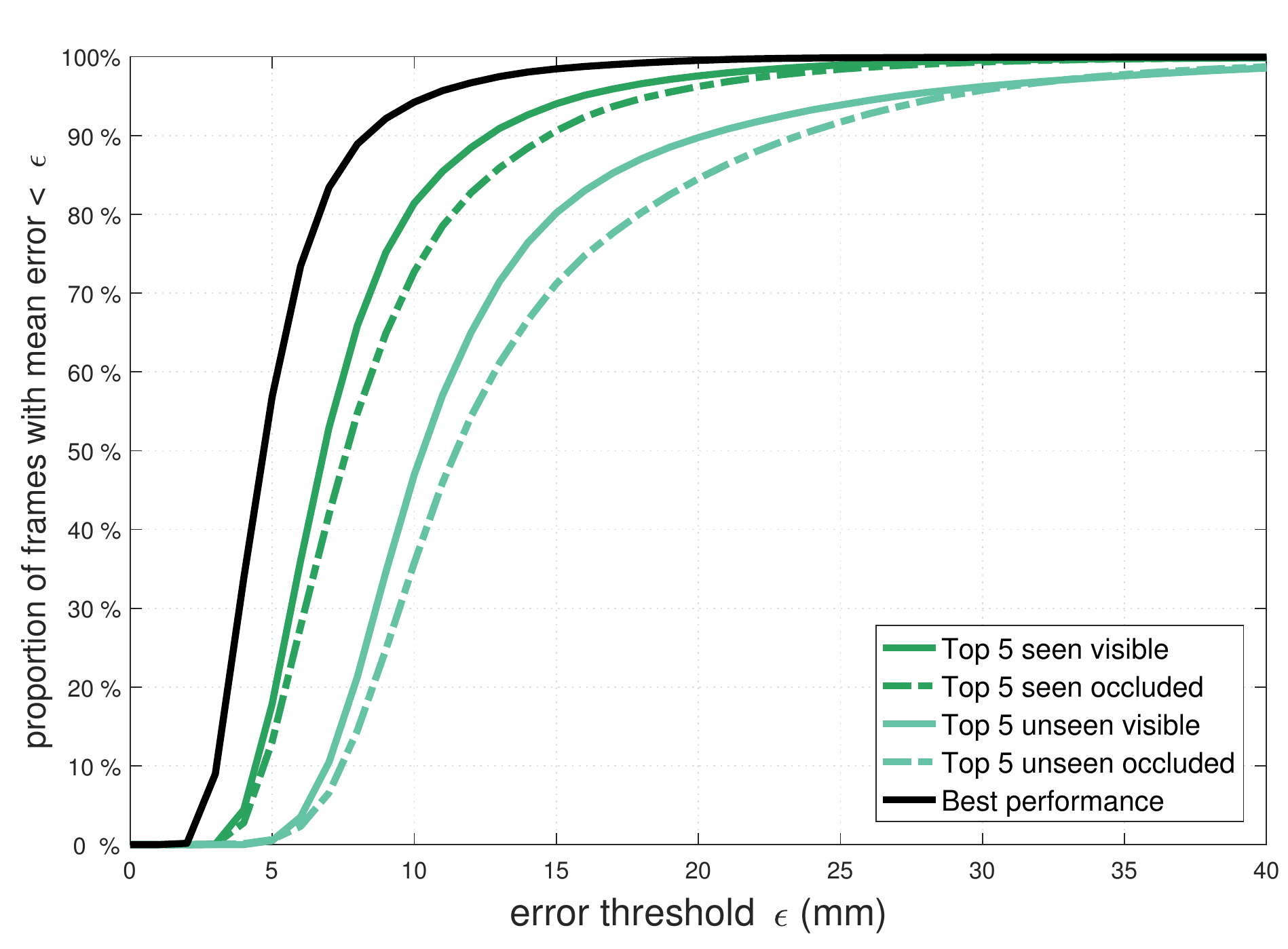} 
        \includegraphics[trim=7.5cm 5cm 11cm 7cm, clip=true,width=0.33\textwidth]{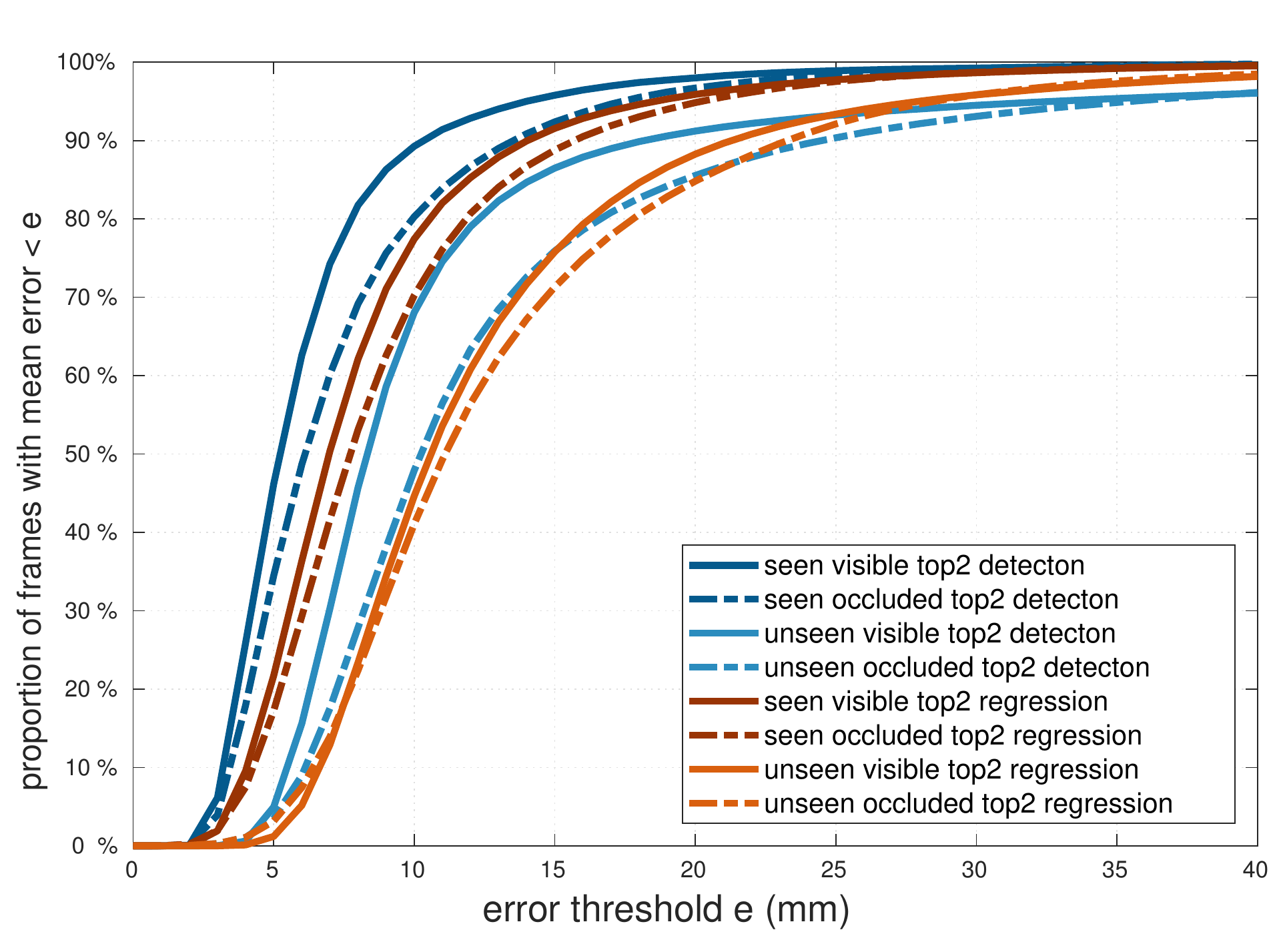}       
        \hfill \\ 
      
		\includegraphics[trim=7.5cm 5cm 11cm 7cm, clip=true,width=0.33\textwidth]{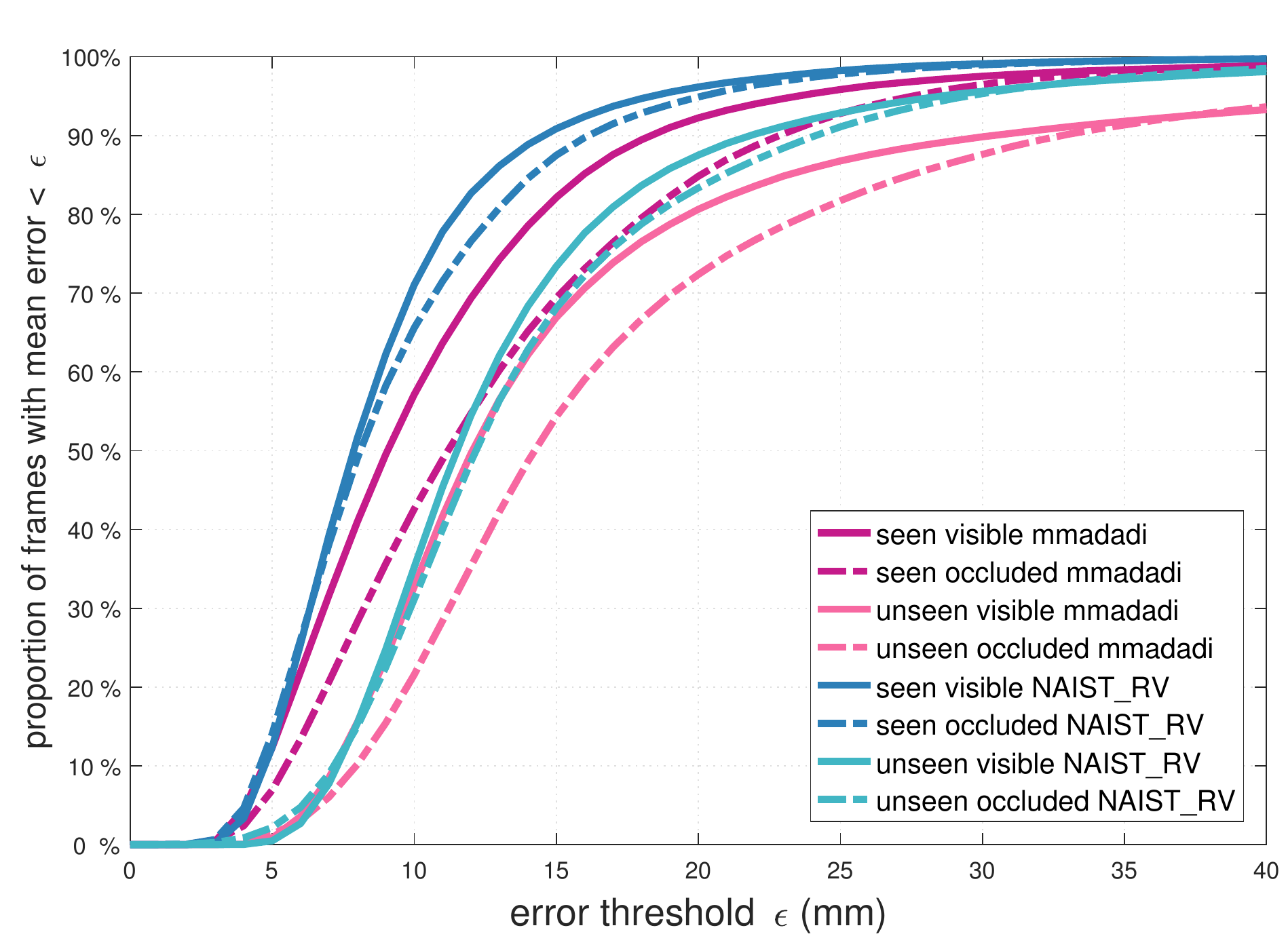}
		\includegraphics[trim=7.5cm 5cm 11cm 7cm, clip=true,width=0.33\textwidth]{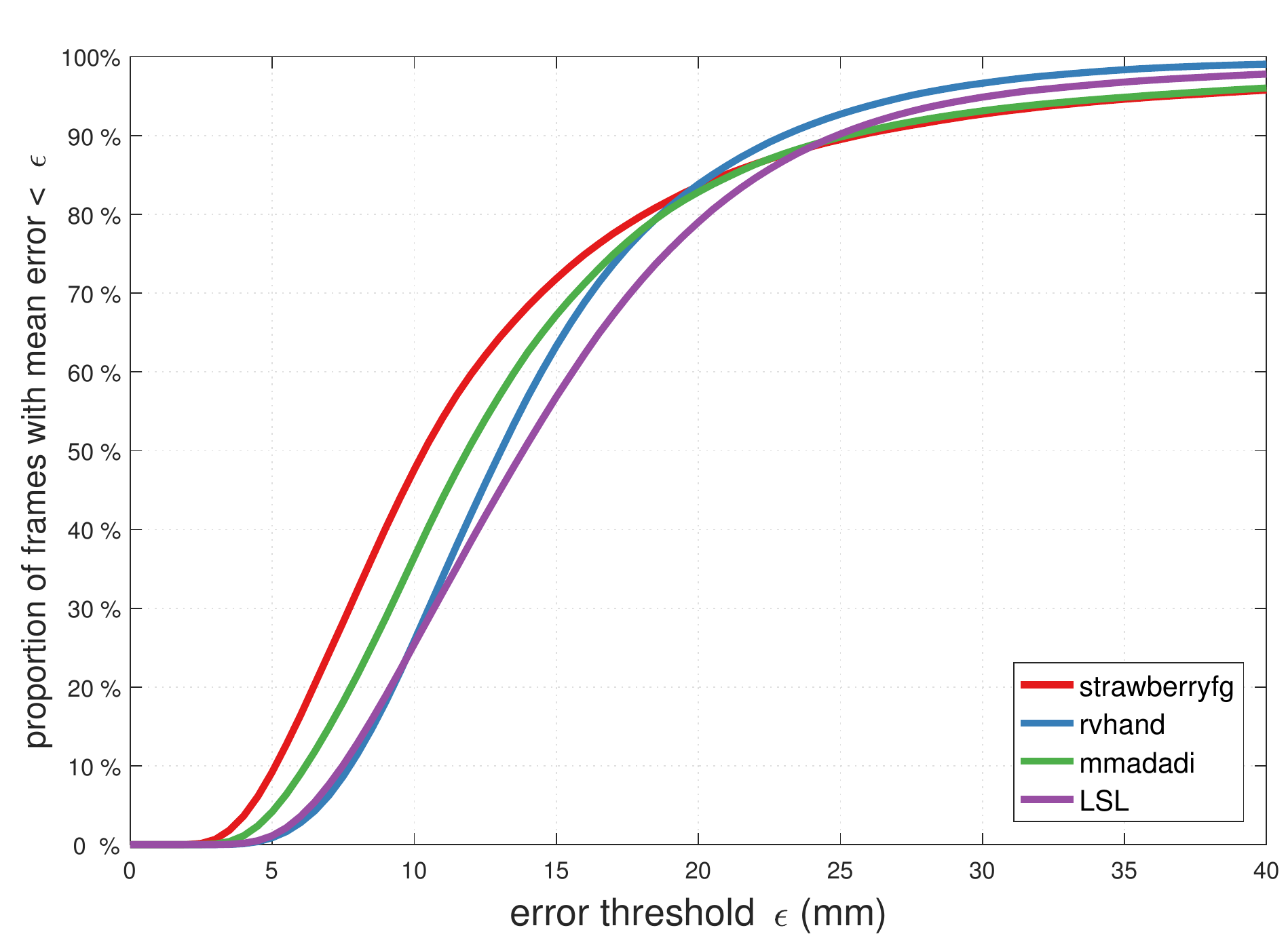}
		\includegraphics[trim=7.5cm 5cm 11cm 7cm, clip=true,width=0.33\textwidth]{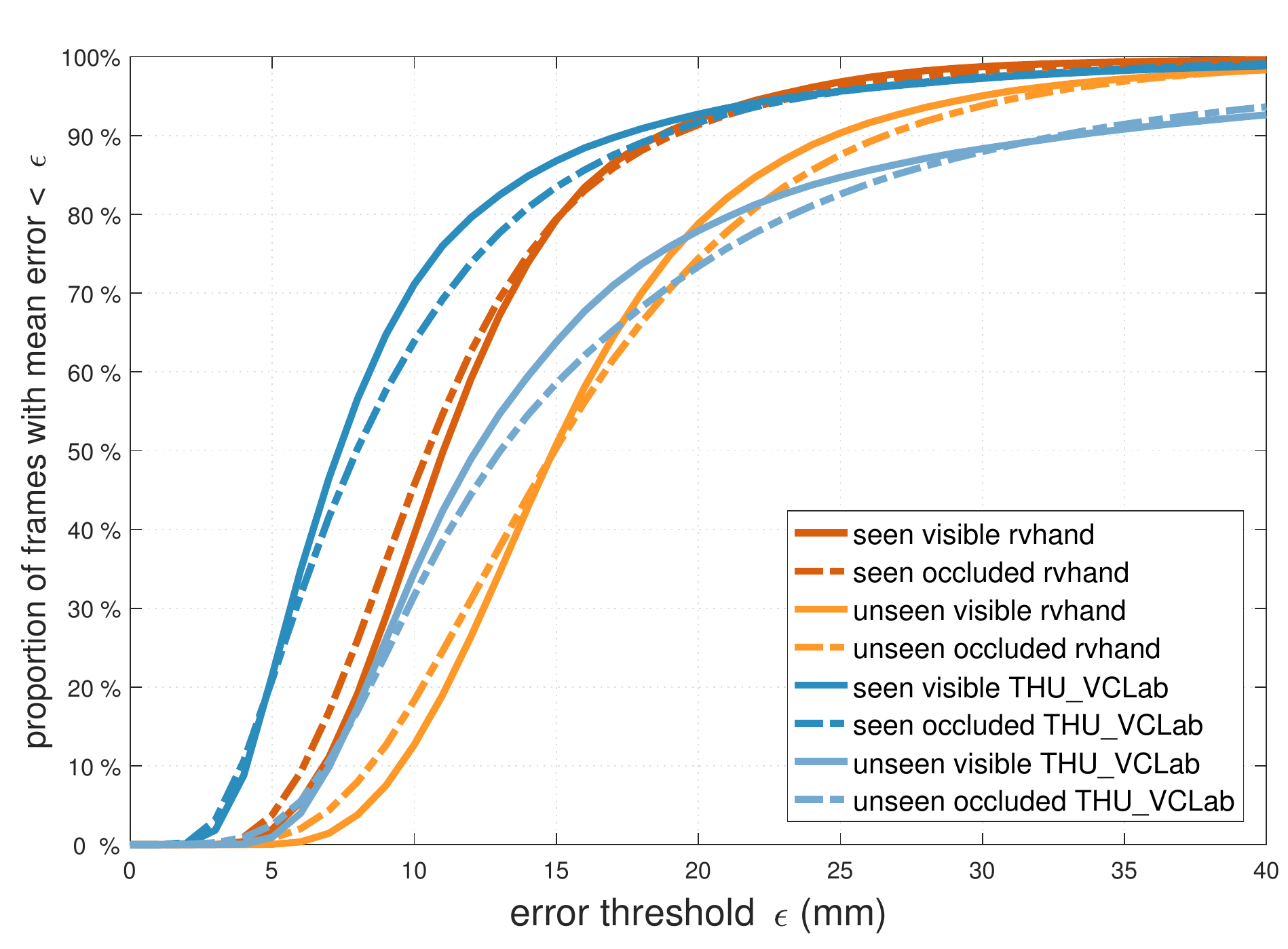}	
		\hfill \\  		
	        
    \caption{{\bf Success rates for different methods.} 
      \textbf{Top-left}: all evaluated methods using all test data. 
      \textbf{Top-middle}: the average of the top five methods for four cases.      
      \textbf{Top-right}: the average of top-two detection-based and regression-based methods in four cases.
      \textbf{Bottom-left}: direct comparison with 2D CNN and 3D CNN, \emph{mmadadi} is a 2D CNN, \emph{NAIST\_RV} has the same structure but replaced 2D CNN with 3D CNN.
      \textbf{Bottom-middle}: comparison among structured methods.
      \textbf{Bottom-right}: comparison of a cascaded multistage method (\emph{THU\_VCLab}) and a one-off method (\emph{rvhand}). Both of them use \emph{REN}~\cite{guo2017towards} as backbone.
      } 
       
	\label{pic:frame_compare_methods}
	\vspace{-5mm} 
\end{center}
\end{figure*}

Figure~\ref{pic:frame_compare_methods} (top-left) shows the success rates based on per-frame average joint errors~\cite{oikonomidis2011efficient} for a varying threshold. The top performer, \emph{V2V-PoseNet}, estimates 70\% of frames with a mean error of less than \SI{10}{\milli\meter}, and 20\% of frames with mean errors under \SI{5}{\milli\meter}. All evaluated methods achieve a success rate greater than 80\% with an average error of less than 20 mm.

As has been noted by \cite{guo2017towards, oberweger2017deepprior}, data augmentation is beneficial, especially for small datasets. 
However, note that even though the top performing methods employ data augmentation, it is still difficult to generalize to hands from unseen subjects, see Table~\ref{tab:avgerrorframe}, with an error gap of around \SI{6}{\milli\meter} between seen and unseen subjects. Some methods generalize better than others, in particular \emph{RCN-3D} is the top performer on unseen subjects, even though it is not the best on seen subjects. 

\vspace{-3mm} 
\subsubsection{Annotation error}
\vspace{-1mm} 
The annotation error takes into account inaccuracies due to small differences of 6D sensor placement for different subjects during the annotation, and uncertainty in the wrist joint location.
To quantify this error we selected  poses for which all methods achieved a maximum error \cite{taylor2012vitruvian} of less than \SI{10}{\milli\meter}.  We denote these as \emph{Easy Poses}. The pose estimation task for these can be considered solved, as shown in Figure~\ref{pic:evalmetric_Uncertain}. The outputs of the top five methods are visually accurate and close to the ground truth. 
We estimate the \emph{Annotation Error} as the error on these poses, which has a mean value of \SI{2.8}{\milli\meter} and a standard deviation of \SI{0.5}{\mm}.

\vspace{-3mm} 
\subsubsection{Analysis by occlusion and unknown subject}
\label{para:upperbound}
\vspace{-1mm} 

\textbf{Average error for four cases}:
To analyze the results with respect to joint visibility and hand shape, we  partition the joints into four groups, based on whether or not they are visible, and whether or not the subject was seen at training time.
Different hand shapes and joint occlusions are responsible for a large proportion of errors, see Table~\ref{tab:avgerrorframe}. The error for unseen subjects is significantly larger than for seen subjects. Moreover, the error for visible joints is smaller than for occluded joints.
Based on the first group (visible, seen), we carry out a best-case performance estimate for the current state-of-the-art. For each frame of seen subjects, we first choose the best result from all  methods, and calculate the success rate based on the average error for each frame, see the black curve in Figure~\ref{pic:frame_compare_methods} (top-middle).

\textbf{2D \vs 3D CNNs}:
We compare two hierarchical methods with similar structure but different representation.
The bottom-left plot of Figure~\ref{pic:frame_compare_methods} shows \emph{mmadadi}~\cite{madadi2017end}, which employs a 2D CNN, and \emph{NAIST\_RV}~\cite{Naisthand17}, using a 3D CNN. \emph{mmadadi} and \emph{NAIST\_RV} have almost the same structure, but \emph{NAIST\_RV}~\cite{Naisthand17} uses a 3D CNN, while \emph{mmadadi}~\cite{madadi2017end} uses a 2D CNN. \emph{NAIST\_RV}~\cite{Naisthand17} outperforms \emph{mmadadi}~\cite{madadi2017end} in all four cases. 

\textbf{Detection-based \vs regression-based methods}:
We compare the average of the top two detection-based methods with the average of the top two regression-based methods. In all four cases, detection-based methods outperform regression-based ones, see the top-right plot of Figure~\ref{pic:frame_compare_methods}. 
In the challenge, the top two methods are detection-based methods, see Table~\ref{tab:methods}. Note that a similar trend can be seen in the field of full human pose estimation, where only one method in a recent challenge was regression-based~\cite{MPIILeader}. 

\textbf{Hierarchical methods}:
Hierarchical constraints can help in the case of occlusion. The hierarchical model in \emph{rvhand}~\cite{akiyama_rvhand17} shows similar performance on visible and occluded joints.
\emph{rvhand}~\cite{akiyama_rvhand17} has better performance on occluded joints when the error threshold is smaller than \SI{15}{\milli\meter}, see the bottom-right plot of Figure~\ref{pic:frame_compare_methods}.
The underlying \emph{REN} module~\cite{guo2017towards}, which includes finger and joint-type constraints seems to be critical. 
Methods using only per-finger constraints, \eg, \emph{mmadadi}~\cite{madadi2017end} and \emph{NAIST\_RV}~\cite{Naisthand17}, generalize less well to occluded joints, see the bottom-left plot of Figure~\ref{pic:frame_compare_methods}.

\textbf{Structural methods:}
We compare four structured methods \emph{LSL}~\cite{LSL17}, \emph{mmadadi}~\cite{madadi2017end}, \emph{rvhand}~\cite{akiyama_rvhand17}, and \emph{strawberryfg}~\cite{Strawberry}, see the bottom-middle plot of Figure~\ref{pic:frame_compare_methods}. \emph{strawberryfg}~\cite{Strawberry} and \emph{mmadadi}~\cite{madadi2017end} have higher success rates when the error threshold is below \SI{15}{\milli\meter}, while \emph{LSL}~\cite{LSL17} and \emph{rvhand}~\cite{akiyama_rvhand17} perform better for thresholds larger than \SI{25}{\milli\meter}. Embedding structural constraints in the loss function has been more successful than including them within the CNN layers. \emph{strawberryfg}~\cite{Strawberry} performs the best, using constraints on phalanges rather than on joints.

\textbf{Single- \vs multi-stage methods}:
Cascaded methods work better than single-stage methods, see the bottom-right plot of Figure~\ref{pic:frame_compare_methods}. Compared to other methods, \emph{rvhand}~\cite{akiyama_rvhand17} and \emph{THU\_VCLab}~\cite{chen2017pose} both embed structural constraints, employing \emph{REN} as their basic structure. \emph{THU\_VCLab}~\cite{chen2017pose} takes a cascaded approach to iteratively update results from previous stages, outperforming \emph{rvhand}~\cite{akiyama_rvhand17}.

\begin{figure}[t]
\begin{center}
		\includegraphics[trim=8cm 1.5cm 10cm 3.2cm, clip=true,width=0.5\textwidth]{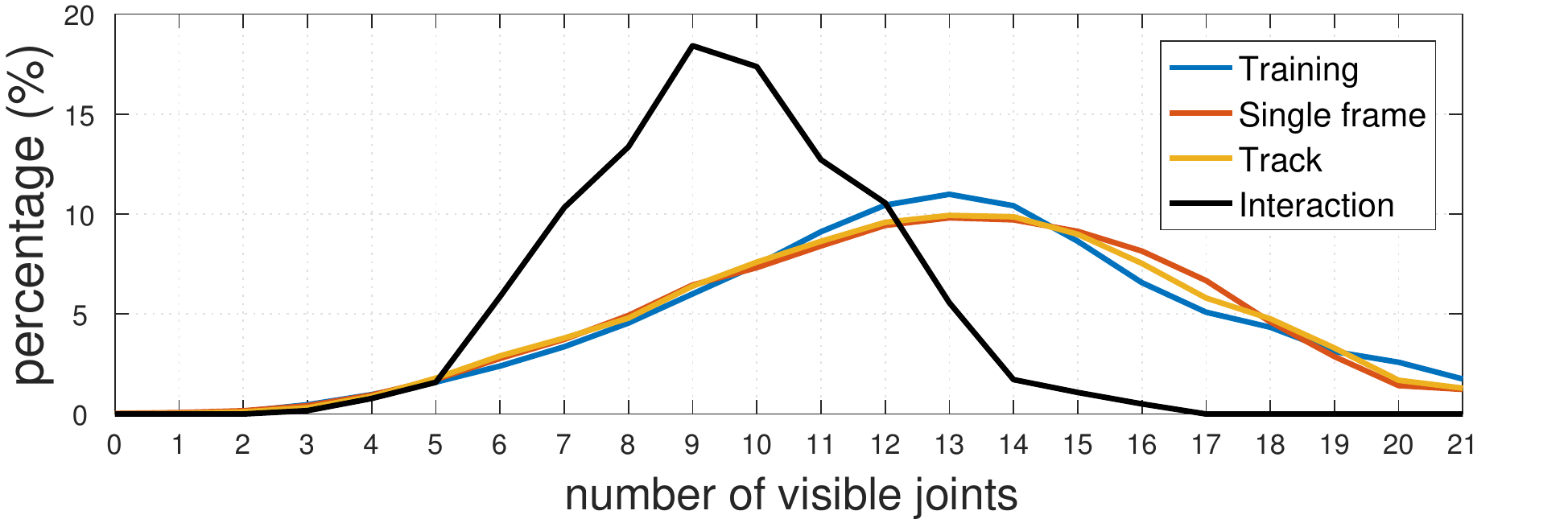}  
		\includegraphics[trim=8cm 1.5cm 10cm 3.2cm, clip=true,width=0.5\textwidth]{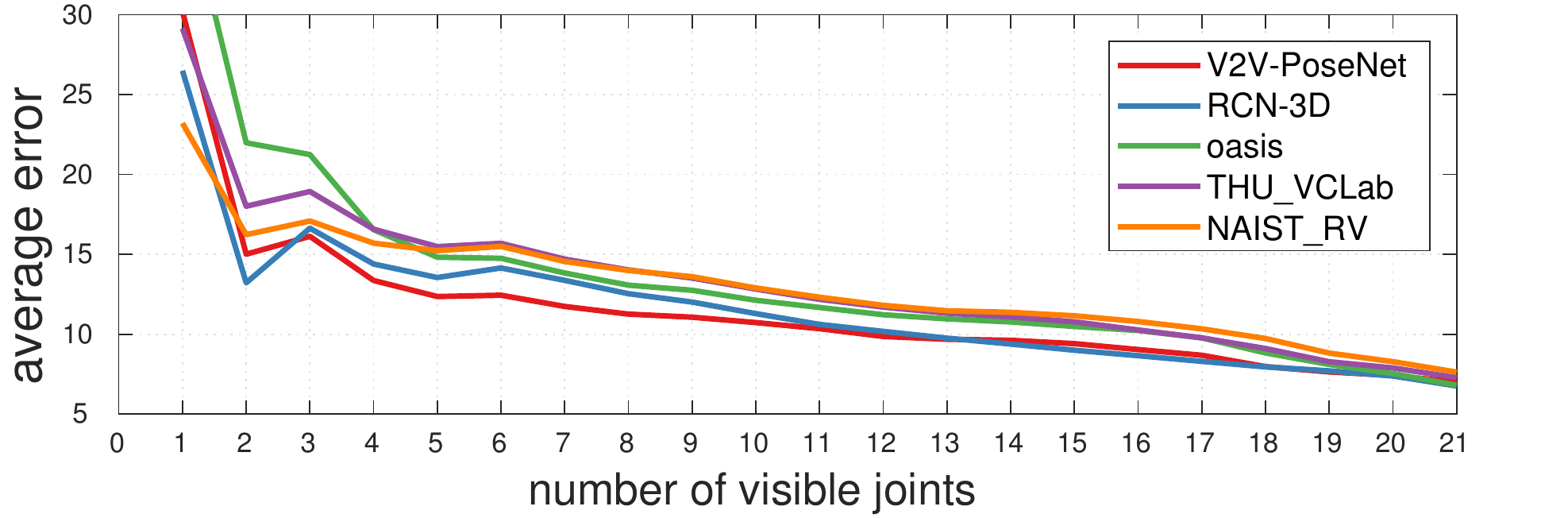}   
		
		\caption{{\bf Joint visibility.} \textbf{Top:} Joint visibility  distributions for training set and testing sets. \textbf{Bottom:} Average error (mm) for different numbers of visible joints and different methods.}        
      
	\label{pic:Challenge_percentage_visible}
		\vspace{-5mm} 
\end{center}
\end{figure}

\vspace{-3mm} 
\subsubsection{Analysis by number of occluded joints}
\vspace{-1mm} 
Most frames contain joint occlusions, see Figure~\ref{pic:Challenge_percentage_visible} (top). 
We assume that a visible joint lies within a  small range of the 3D point cloud. We therefore detect joint occlusion by thresholding the distance between the joint's depth annotation value and its re-projected depth value.  As shown in Figure~\ref{pic:Challenge_percentage_visible} (bottom), the average error decreases nearly monotonously for increasing numbers of visible joints.

\begin{figure}[t]
\begin{center}
		\includegraphics[trim=7.5cm 0.4cm 9cm 2cm, clip=true,width=0.5\textwidth]{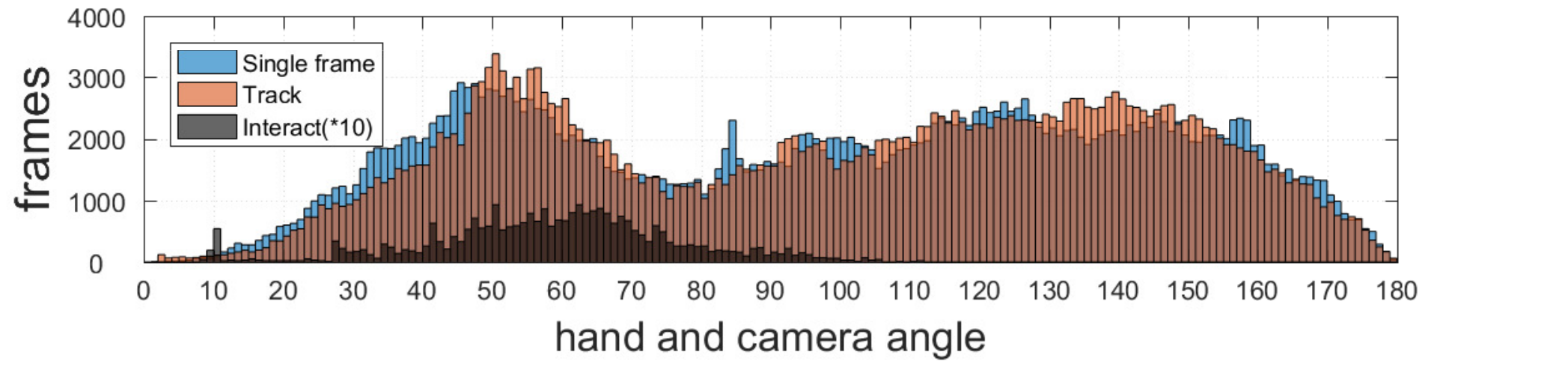} 
		\includegraphics[trim=7.5cm 1.4cm 9cm 3.2cm, clip=true,width=0.5\textwidth]{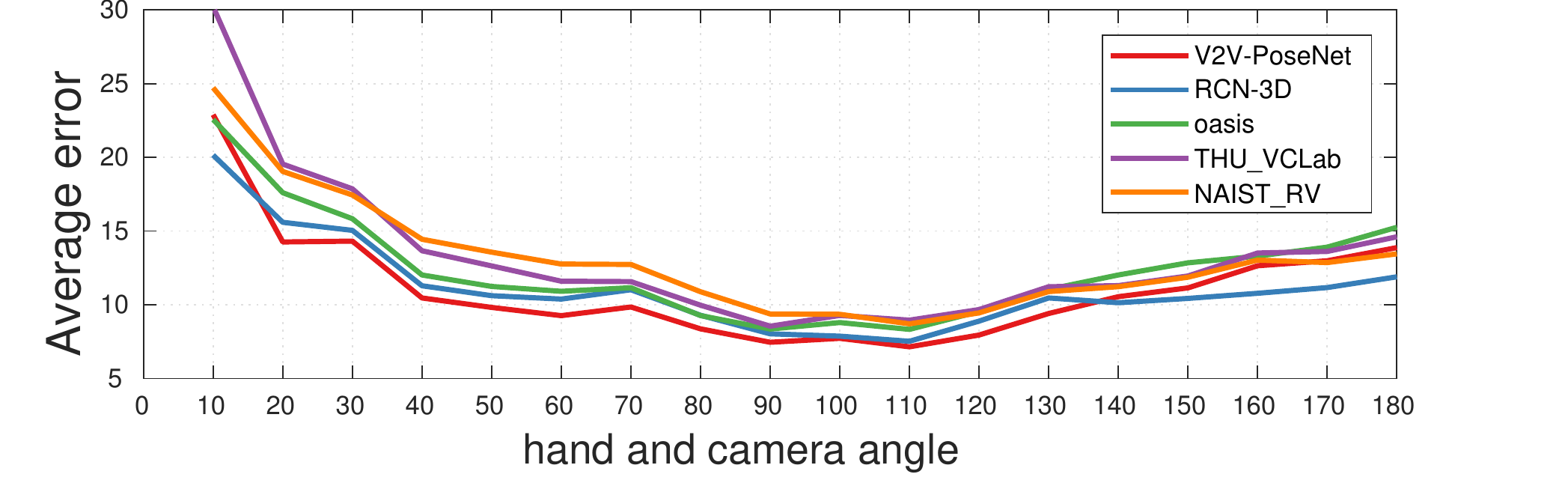}               
        
    \caption{{\bf View point distributions.} The error is significantly higher for small angles between hand and camera orientations.}        
	\label{pic:Challenge_Perpendicular}
		\vspace{-5mm} 
\end{center}
\end{figure}

\begin{figure}[t]
\begin{center}
		\includegraphics[trim=7.5cm 0.4cm 9cm 2cm, clip=true,width=0.5\textwidth]{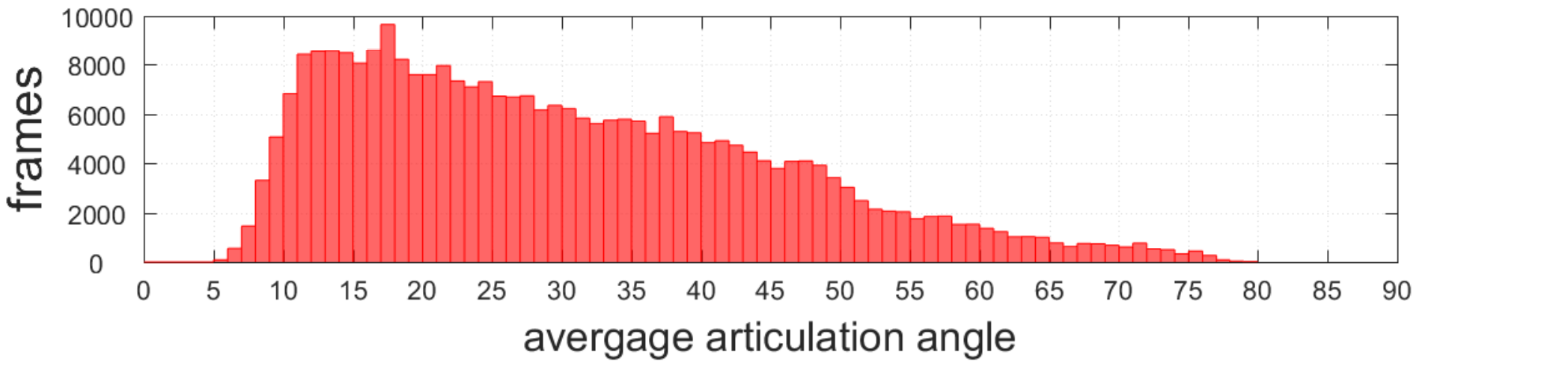} 
		\includegraphics[trim=7.5cm 1.4cm 9cm 3.2cm, clip=true,width=0.5\textwidth]{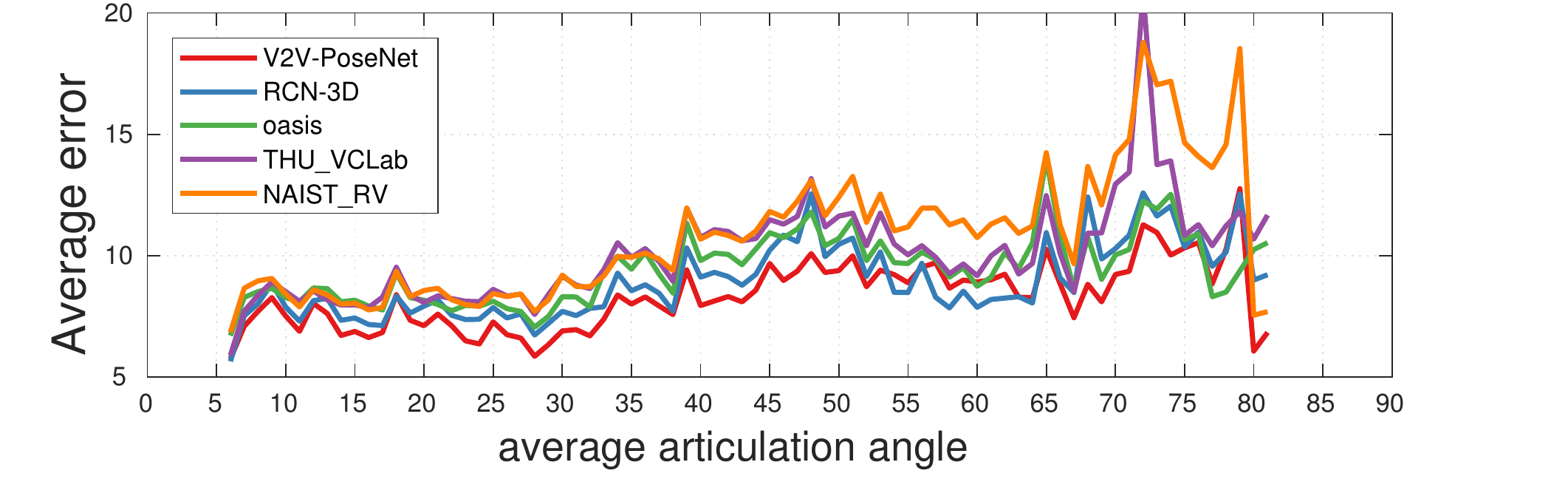}   
    \caption{{\bf Articulation distribution.} Errors increase for larger finger articulation angles.}        
	\label{pic:Challenge_articulation} 
		\vspace{-5mm} 
\end{center}
\end{figure}

\vspace{-3mm} 
\subsubsection{Analysis based on view point}
\vspace{-1mm} 
The view point is defined as the angle between the palm and camera directions. The test data covers a wide range of view points for the \emph{Single frame pose estimation} task, see Figure~\ref{pic:Challenge_Perpendicular} (top). View points in the [70, 120] range have a low mean error of below \SI{10}{\milli\meter}. View points in the [0, 10] range have a significantly larger error due to the amount of self occlusion. View points in the [10, 30] range have an average error of \SI{15}-\SI{20}{\milli\meter}. View point ranges of [30,70] and [120, 180] show errors of \SI{10}-\SI{15}{\milli\meter}. Third-person and egocentric views are typically defined by the hand facing toward or away from the camera, respectively. However, as shown in Figure~\ref{pic:Challenge_Perpendicular}, there is no clear separation by view point, suggesting a uniform treatment of both cases is sensible. 
Note that \emph{RCN-3D}~\cite{molchanov_Nvidia17} outperforms others with a margin of 2-\SI{3}{\milli\meter} on extreme view points in the range of [150,180] degrees due to their depth prediction stage.

\begin{figure*}[t]
\begin{center}
		\includegraphics[trim=7.5cm 5cm 11cm 7cm, clip=true,width=0.33\textwidth]{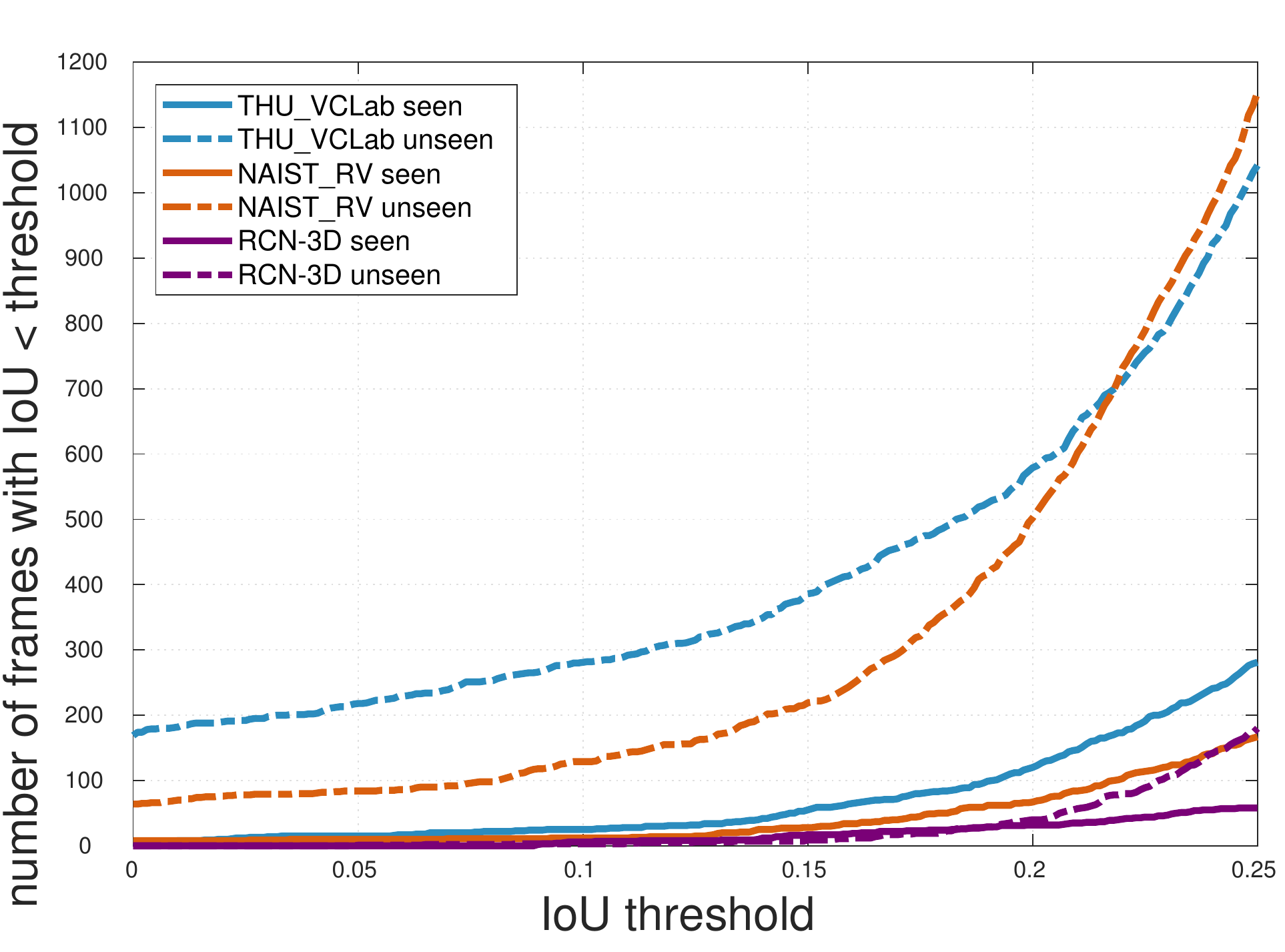}  
        \includegraphics[trim=7.5cm 5cm 11cm 7cm, clip=true,width=0.33\textwidth]{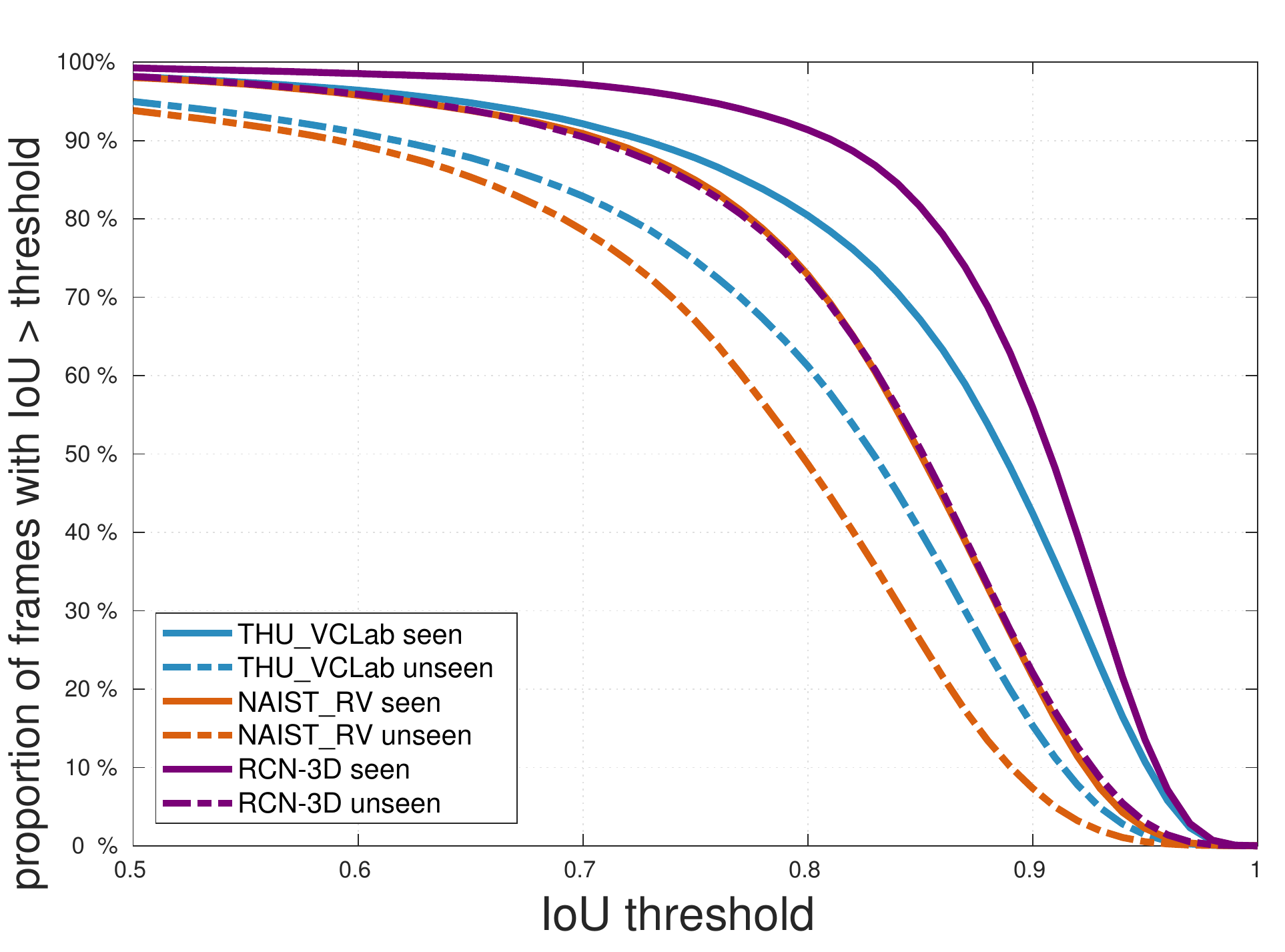}
        \includegraphics[trim=7.5cm 5cm 11cm 7cm, clip=true,width=0.33\textwidth]{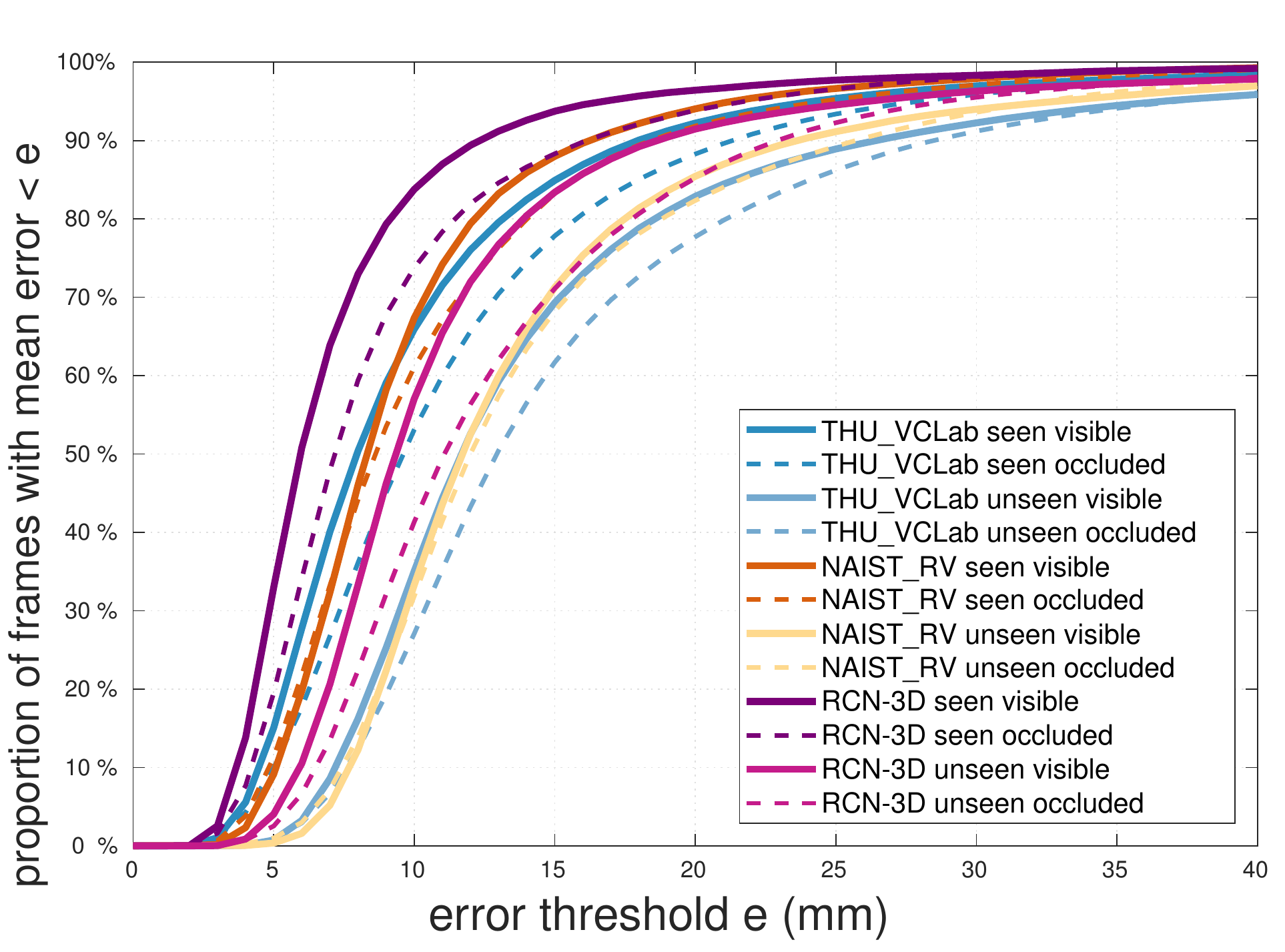}
                      
        \hfill \\
        \vspace{-3mm} 
        
    \caption{{\bf Error curves for hand tracking.} \textbf{Left}: number of frames with IoU below threshold. \textbf{Middle}: success rate for the detection part of each method in two cases. \textbf{Right}: success rate for three methods in four cases.}    
	\label{pic:tracking_compare_curve}
	\vspace{-3mm} 
\end{center}
\end{figure*}

\begin{figure*}[t]
\begin{center}
        \includegraphics[trim=7.5cm 5cm 11cm 7cm, clip=true,width=0.33\textwidth]{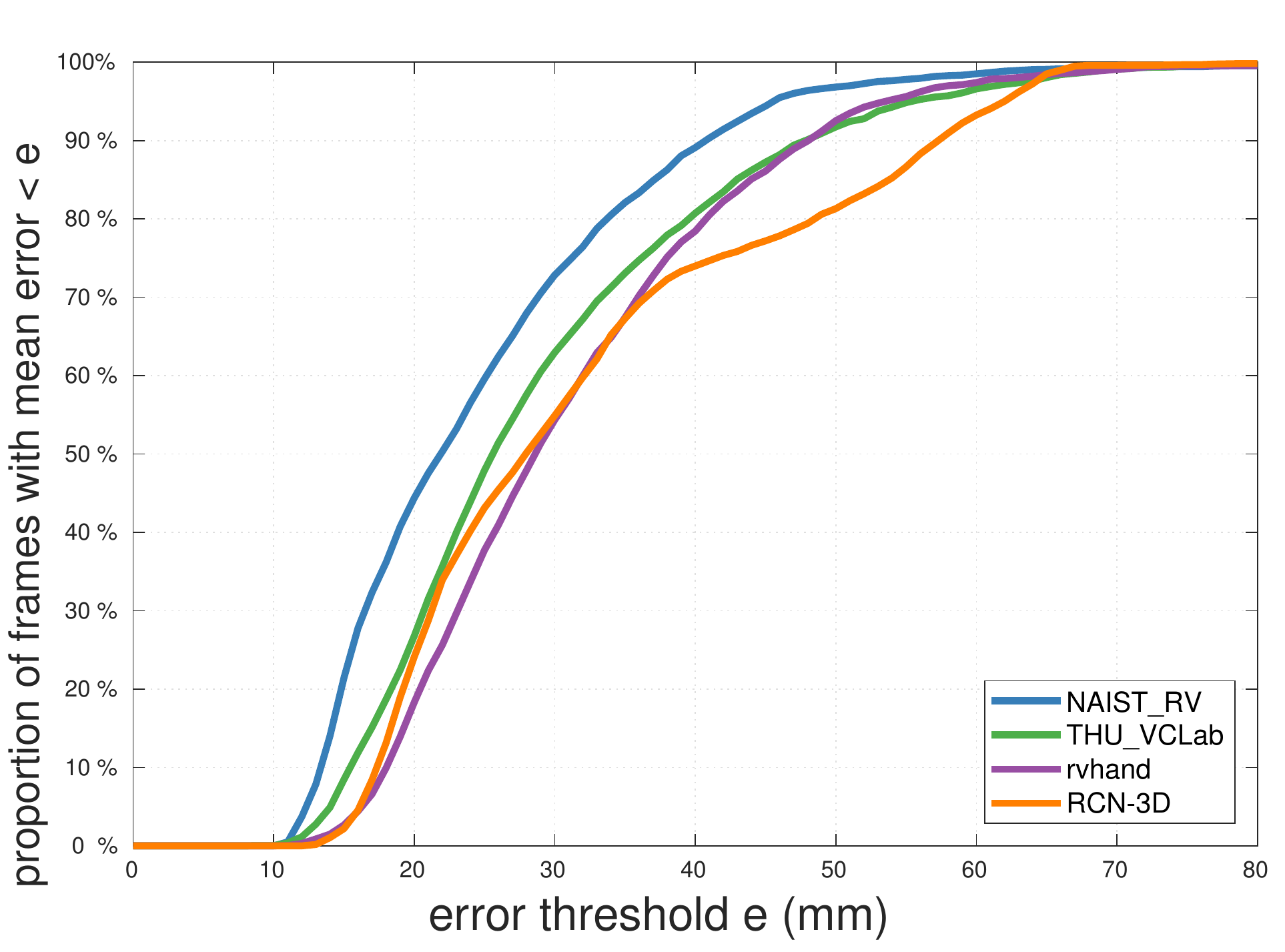}
        \includegraphics[trim=7.5cm 5cm 11cm 7cm, clip=true,width=0.33\textwidth]{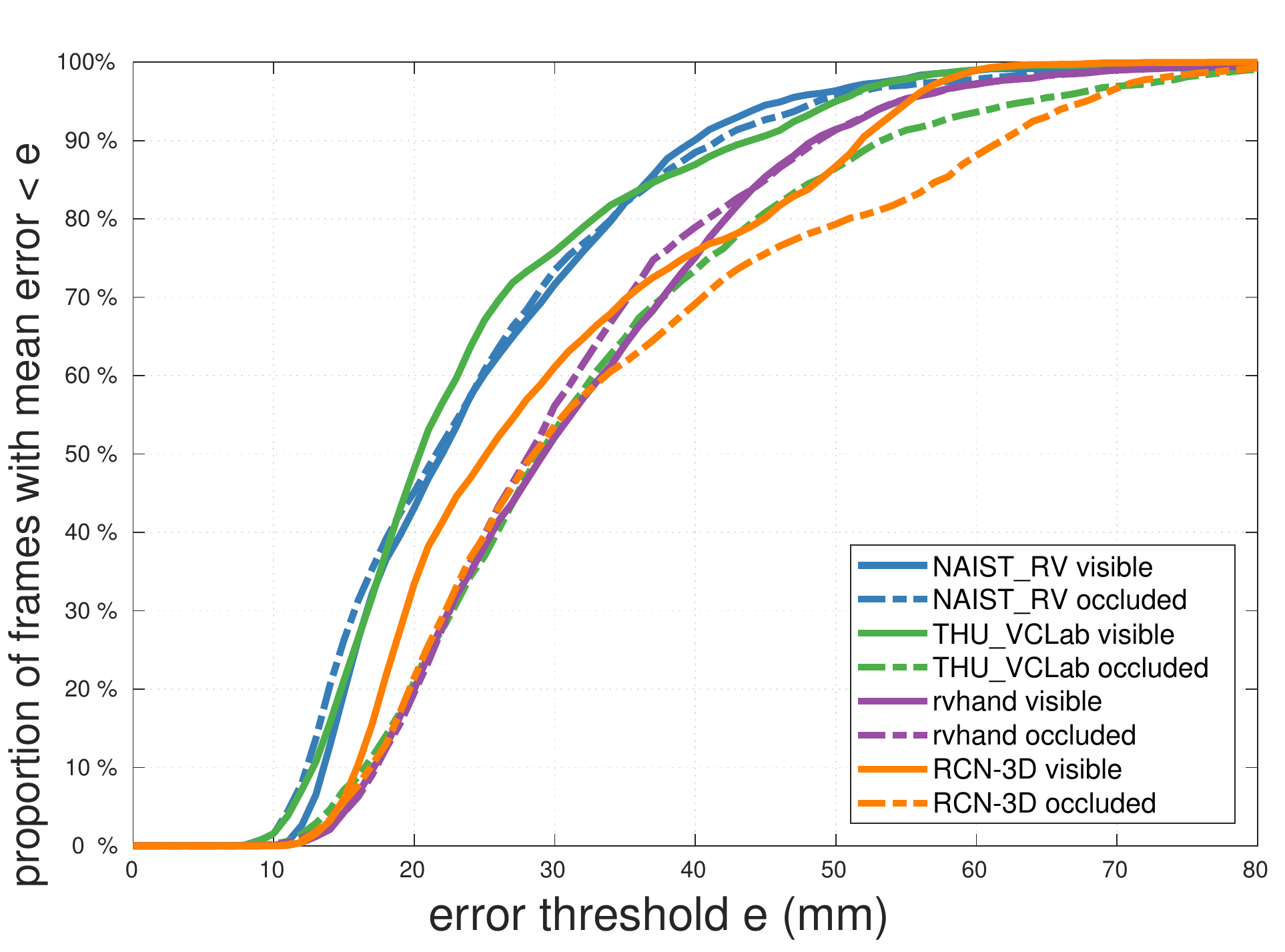}   
        \includegraphics[trim=7.5cm 5cm 11cm 7cm, clip=true,width=0.33\textwidth]{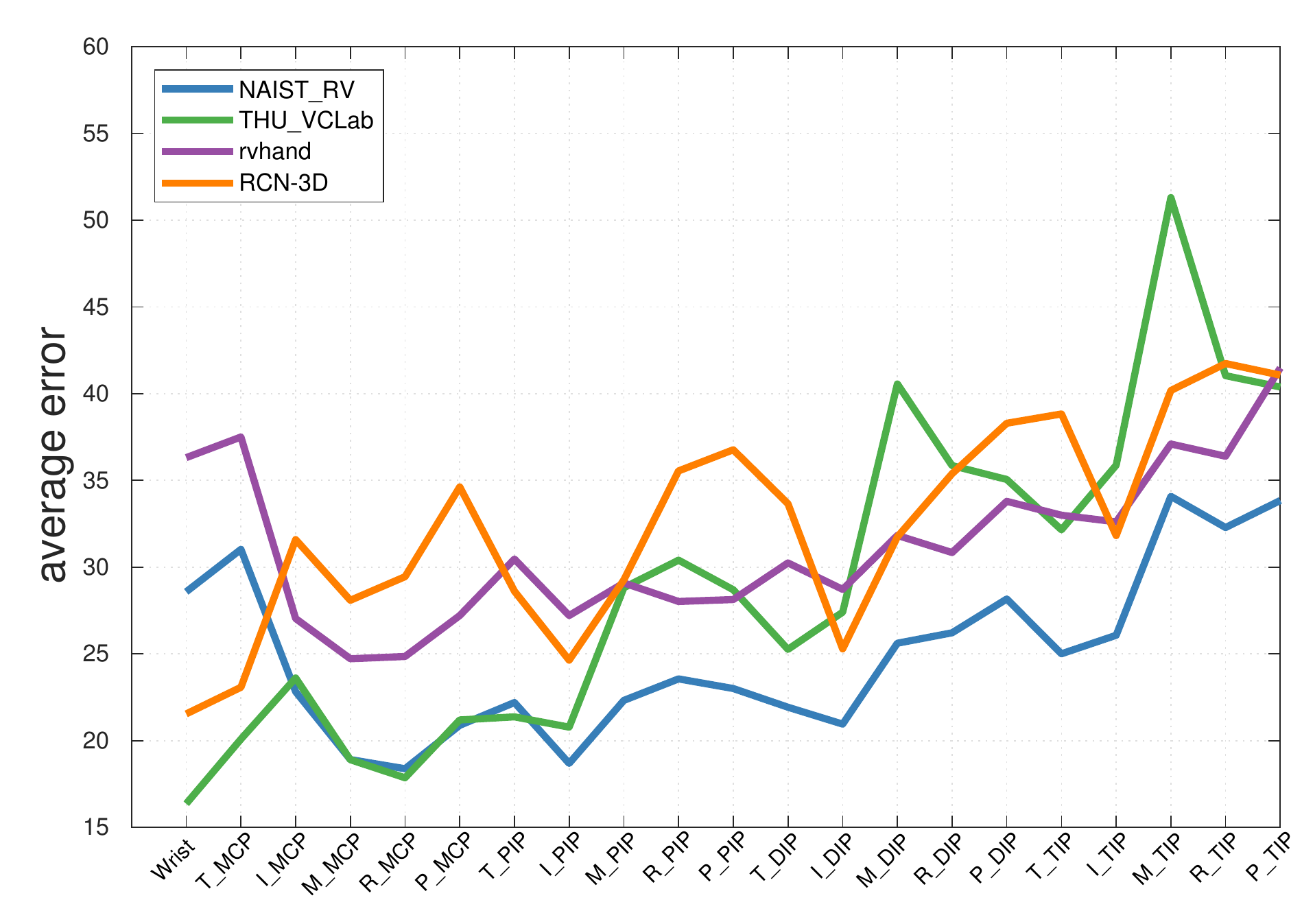} 	
               
                \hfill \\   
                \vspace{-3mm}      
    \caption{{\bf Error curves for hand-object interaction.} \textbf{Left}: success rate for each method using average error per-frame. \textbf{Middle}: success rate for visible and occluded joints. \textbf{Right}: average error for each joint.
    }    
	\label{pic:interaction_compare_curve}
	\vspace{-5mm} 
\end{center}
\end{figure*}

\begin{figure}[t]
\begin{center}
        \includegraphics[trim=8cm 1.5cm 10cm 3.2cm, clip=true,width=0.5\textwidth]{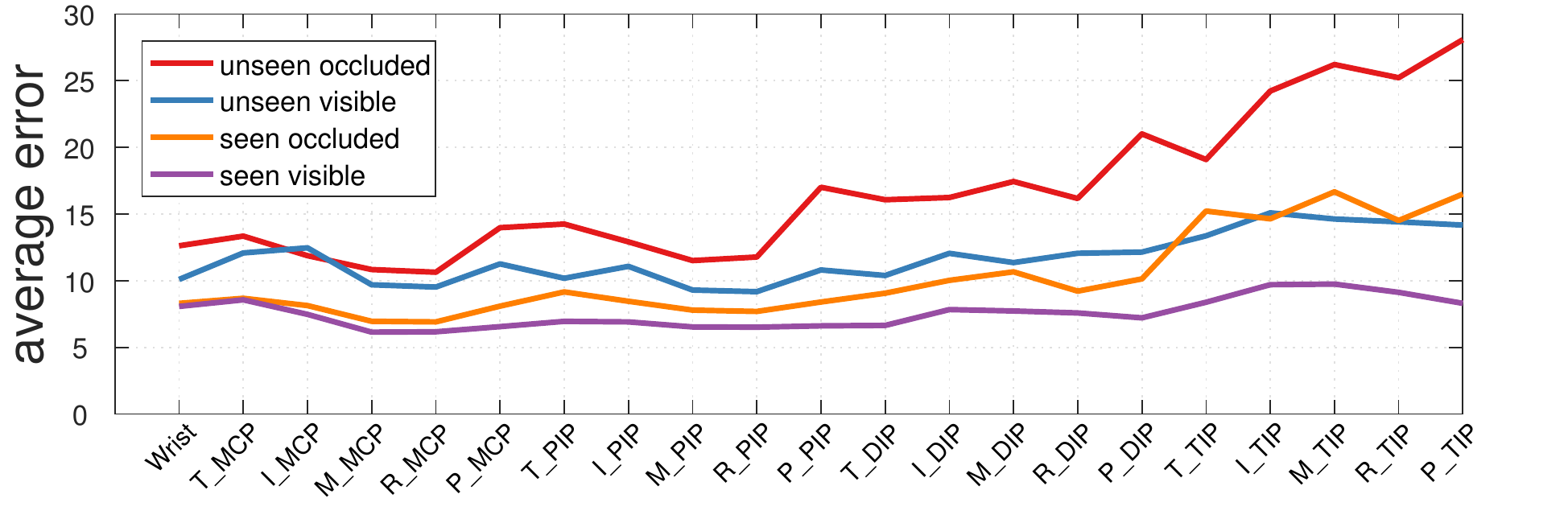}  

    \caption{{\bf Average error of the top five methods for each joint in the \emph{Single frame pose estimation} task.} Finger tips have larger errors than other joints. For non-tip joints, joints on ring finger and middle finger have lower average errors than other fingers. `T', `I', `M', `R', `P' denotes `Thumb', `Index', `Middle', `Ring', and `Pinky' finger, respectively.}
     
	\label{pic:Challenge_jointnames}
		\vspace{-10mm} 
\end{center}
\end{figure} 
    
\vspace{-3mm} 
\subsubsection{Analysis based on articulation}
\vspace{-1mm} 
We evaluate the effect of hand articulation on estimation accuracy, measured as the average of 15 finger flexion angles, see Figure~\ref{pic:Challenge_articulation}. To reduce the influence from other factors such as view point, we select frames with view point angles within the range of [70, 120]. We evaluate the top five performers, see Figure~\ref{pic:Challenge_articulation} (bottom). For articulation angles smaller than 30 degrees, the mean error is \SI{7}{\milli\meter}, when the average articulation angle increases to the range of [35, 70], errors increase to 9-10 mm. When the articulation angle is larger than 70 degrees, close to a fist pose, the mean error increases to over \SI{12}{\milli\meter}. 
\vspace{-3mm} 
\subsubsection{Analysis by joint type}
\vspace{-1mm} 
As before we group joints according to their visibility and the presence of the subject in the training set.
We report the top five performers, see Figure~\ref{pic:Challenge_jointnames}. 
For the easiest case (visible joints of seen subjects), all 21 joints have a similar average error of \SI{6}-\SI{10}{\milli\meter}. 
For seen subjects, along the kinematic hand structure from the wrist to finger tips,  occluded joints have increasingly larger errors, reaching \SI{14}{\milli\meter} in the finger tips. 
Visible joints of unseen subjects have larger errors (10-\SI{13}{\milli\meter}) than that of seen subjects. Occluded joints of unseen subjects have the largest errors, with a relatively smaller error for the palm, and larger errors for finger tips (\SI{24}-\SI{27}{\milli\meter}). We draw two conclusions: (1)~all the top performers have difficulty in generalizing to hands from unseen subjects, (2)~occlusions have more effect on finger tips than other joints. 
An interesting observation is that middle and ring fingers tend to have smaller errors in \emph{MCP} and \emph{PIP} joints than other fingers. One reason may be that the motion of these fingers is more restricted. 
The thumb's \emph {MCP} joint has a larger error than for other fingers, because it tends to have more discrepancy among different subjects.

\begin{table}[t!]
\small
  \centering
  \resizebox{\columnwidth}{!}{
  \begin{tabular}{lL{5cm}L{1cm}}
  \toprule 
  \bf Method   & \bf Model   & \bf AVG  \\ 
  \midrule
\emph{RCN-3D\_track}~\cite{molchanov_Nvidia17}
& scanning window + post-processing + pose estimator~\cite{molchanov_Nvidia17} 
& 10.5
\\
\hline
\emph{NAIST\_RV\_track}~\cite{Naisthand17}
& hand detector~\cite{ronneberger2015u}
+ hand verifier  
+ pose estimator~\cite{Naisthand17}  
&12.6 
\\
\hline
\emph{THU\_VCLab\_track}~\cite{chen2017pose} 
& Estimation~\cite{chen2017pose} with the aid of tracking + re-initialization \cite{ren2015faster} 
&13.7
\\

  \bottomrule
  \end{tabular}}
    \caption{\textbf{Methods evaluated on 3D hand pose tracking.} The last column is the average error in mm for all frames.}
  \label{tab:methods_tracking} 
  \vspace{-5mm} 
\end{table}

\vspace{-2mm} 
\subsection{Hand pose tracking}
\vspace{-1mm} 

In this task we evaluate three state-of-the-art methods, see Table~\ref{tab:methods_tracking} and Figure~\ref{pic:tracking_compare_curve}. 
Discriminative methods~\cite{chen2017pose,molchanov_Nvidia17, Naisthand17} break tracking into two sub-tasks: detection and hand pose estimation, sometimes merging the sub-tasks~\cite{chen2017pose}. 
Based on the detection methods, 3D hand pose estimation can be grouped into pure tracking~\cite{molchanov_Nvidia17}, tracking-by-detection~\cite{Naisthand17}, and a combination of tracking and re-initialization~\cite{chen2017pose}, see Table~\ref{tab:methods_tracking}.

\textbf{Pure tracking:}
\emph{RCN-3D\_track} estimate the bounding box location by scanning windows based on the result in the previous frame, including a motion estimate. Hand pose within the bounding box is estimated using \emph{RCN-3D}~\cite{molchanov_Nvidia17}. 

\textbf{Tracking-by-detection:}
\emph{NAIST\_RV\_track} is a tracking-by-detection method with three components: hand detector, hand verifier, and pose estimator. The hand detector is built on U-net~\cite{ronneberger2015u} to predict a binary hand-mask, which, after verification, is passed to the pose estimator \emph{NAIST\_RV}~\cite{Naisthand17}. If verification fails, the result from the previous frame is chosen.

\textbf{Hybrid tracking and detection:}
\emph{THU\_VCLab\_track}~\cite{chen2017pose} makes use of the previous tracking result and the current frame's scanning window. The hand pose of the previous frame is used as a guide to predict the hand pose in the current frame. 
The previous frame's bounding box is used for the current frame. During fast hand motion, Faster R-CNN~\cite{ren2015faster} is used for re-initialization.

\textbf{Detection accuracy:}
We first evaluate the detection accuracy by evaluating the bounding box overlap, \ie, the intersection over union (IoU) of the detection and ground truth bounding boxes, see Figure~\ref{pic:tracking_compare_curve} (middle). Overall, \emph{RCN-3D\_track}  is more accurate than \emph{THU\_VCLab\_track}, which itself outperforms \emph{NAIST\_RV\_track}. Pure detection methods have a larger number of false negatives, especially when multiple hands appear in the scene, see Figure~\ref{pic:tracking_compare_curve} (left).  There are 72 and 174 missed detections (IoU of zero), for \emph{NAIST\_RV\_track} and \emph{THU\_VCLab\_track}, respectively. By tracking and re-initializing, \emph{THU\_VCLab\_track}  achieves better detection accuracy overall. \emph{RCN-3D\_track}, using motion estimation and single-frame hand pose estimation, shows the lowest error.

\textbf{Tracking accuracy} is shown in Figure~\ref{pic:tracking_compare_curve} (right). Even through \emph{THU\_VCLab} performs better than \emph{NAIST\_RV} in the \emph{Single frame pose estimation} task, \emph{NAIST\_RV\_track} performs better on the tracking tasks due to per-frame hand detection.

\begin{table}[t!]
\small
  \centering
  \resizebox{\columnwidth}{!}{
  \begin{tabular}{lL{5cm}L{1cm}}
  \toprule 
  \bf Method   & \bf Model  & \bf AVG \\ 
  \midrule   

NAIST\_RV\_obj \cite{Naisthand17}
& Hand-object segmentation CNN + pose estimation~\cite{Naisthand17} 
&25.0
\\
\hline

THU\_VCLab\_obj \cite{chen2017pose}
& Hand-object segmentation (intuitive) + pose estimation~\cite{chen2017pose}
&29.2
\\
\hline

rvhand\_obj \cite{akiyama_rvhand17} 
& Hand-object segmentation CNN + pose estimation~\cite{akiyama_rvhand17} 
&31.3
\\
\hline

RCN-3D\_obj \cite{molchanov_Nvidia17}
& Two RCNs: Feature maps of first are used in the second RCN's stage 2.
& 32.4
\\

  \bottomrule
  \end{tabular}}
    \caption{\textbf{Methods evaluated on hand pose estimation during hand-object interaction.} The last column is the average error (mm) for all frames. }
  \label{tab:methods_interaction} 
  \vspace{-5mm} 
\end{table}

\subsection{Hand object interaction}
For this task, we evaluate four state-of-the-art methods, see Table~\ref{tab:methods_interaction} and Figure~\ref{pic:interaction_compare_curve}. Compared to the other two tasks there is significantly more occlusion, see Figure~\ref{pic:Challenge_percentage_visible} (top). 
Methods explicitly handling occlusion achieve higher accuracy with errors in the range of \SI{25}-\SI{29}{\milli\meter}: (1) \emph{NAIST\_RV\_obj}~\cite{Naisthand17} and rvhand\_obj~\cite{akiyama_rvhand17} segment the hand area from the object using a network. 
(2) \emph{THU\_VCLab\_obj}~\cite{chen2017pose} removes the object region from cropped hand images with image processing operations \cite{serra1982image}. (3) \emph{RCN-3D\_obj}~\cite{molchanov_Nvidia17} modify their original network to infer the depth values of 2D keypoint locations.

Current state-of-the-art methods have difficulty generalizing to the hand-object interaction scenario. However, \emph{NAIST\_RV\_obj}~\cite{Naisthand17} and rvhand\_obj~\cite{akiyama_rvhand17} show similar performance for visible joints and occluded joints, indicating that CNN-based segmentation can better preserve structure than image processing operations, see the middle plot of Figure~\ref{pic:interaction_compare_curve}.

\vspace{0mm} 
\section{Discussion and conclusions}
\vspace{-1mm} 

The analysis of the top 10 among 17 participating methods from the \textit{HIM2017} challenge~\cite{HIM2017} provides insights into the current state of 3D hand pose estimation. 

(1)~3D volumetric representations used with a 3D CNN show high performance, possibly by better capturing the spatial structure of the input depth data.

(2)~Detection-based methods tend to outperform regression-based methods, however, regression-based methods can achieve good performance using explicit spatial constraints. Making use of richer spatial models, \eg, bone structure~\cite{sun2017compositional}, helps further. Regression-based methods perform better in extreme view point cases~\cite{molchanov_Nvidia17}, where severe occlusion occurs.

(3)~While joint occlusions pose a challenge for most methods, explicit modeling of structure constraints and spatial relation between joints can significantly narrow the gap between errors on visible and occluded joints~\cite{akiyama_rvhand17}.  

(4)~Discriminative methods still generalize poorly to unseen hand shapes. Data augmentation and scale estimation methods model only global shape changes, but not local variations. Integrating hand models with better generative capability may be a promising direction.

(5)~Isolated 3D hand pose estimation achieves low mean errors (\SI{10}{\milli\meter}) in the view point range of [70, 120] degrees. However, errors remain large for extreme view points, \eg, view point range of [0,10], where the hand is facing away from the camera. Multi-stage methods~\cite{molchanov_Nvidia17} tend to perform better in these cases.

(6)~In hand tracking, current discriminative methods divide the problem into two sub-tasks: detection and pose estimation, without using the hand shape provided in the first frame. Hybrid methods may work better by using the provided hand shape.

(7)~Current methods perform well on single hand pose estimation when trained on a million-scale dataset, but have difficulty in generalizing to hand-object interaction. Two directions seem promising, (a)~designing better hand segmentation methods, and (b)~training the model with large datasets containing hand-object interaction.

\vspace{1mm}
\textbf{Acknowledgement:} This work was partially supported by Huawei Technologies.

{\small
\bibliographystyle{ieee}
\bibliography{egbib}
}

\end{document}